\documentclass{article} 
\usepackage{iclr2023_conference,times}

\usepackage{amsmath,amsfonts,bm}

\def\eqref#1{equation~\ref{#1}}

\def\1{\bm{1}}

\DeclareMathAlphabet{\mathsfit}{\encodingdefault}{\sfdefault}{m}{sl}
\SetMathAlphabet{\mathsfit}{bold}{\encodingdefault}{\sfdefault}{bx}{n}

\usepackage{hyperref}
\usepackage{url}
\usepackage{enumitem}
\usepackage{amsmath,amssymb}
\usepackage{graphicx}   
\usepackage{xcolor}
\usepackage{bm}
\usepackage{multirow}
\usepackage{adjustbox}
\usepackage{wrapfig}

\newcommand{\eg}{\textit{e.g.},}
\newcommand{\ie}{\textit{i.e.},}

\title{CircNet: Meshing 3D Point Clouds with \\Circumcenter Detection}

\begin{document}

\author{Huan Lei,~~Ruitao Leng,~~Liang Zheng,~~Hongdong Li
\\
School of Computing, The Australian National University \\
}

%

\newcommand{\fix}{\marginpar{FIX}}
\newcommand{\new}{\marginpar{NEW}}

\iclrfinalcopy 

\maketitle

\begin{abstract}
Reconstructing 3D point clouds into triangle meshes is a key problem in computational geometry and surface reconstruction. Point cloud triangulation solves this problem by providing edge information to the input points. Since no vertex interpolation is involved, it is beneficial to preserve 
sharp details on the surface.
Taking advantage of learning-based techniques in triangulation, existing methods enumerate the complete combinations of candidate triangles, 
which is both complex and inefficient.
In this paper, we leverage the \textit{duality} between a triangle and its circumcenter, and introduce a deep neural network that detects the circumcenters to achieve point cloud triangulation. Specifically, we introduce multiple anchor priors to divide the neighborhood space of each point. The neural network then learns to predict the presences and locations of circumcenters under the guidance of those anchors. 
We extract the triangles dual to the detected circumcenters to form a primitive mesh, from which an edge-manifold mesh is produced via simple post-processing. Unlike existing learning-based triangulation methods, the proposed method bypasses an exhaustive enumeration of triangle combinations and local surface parameterization. 
We validate the efficiency, generalization, and robustness of our method on prominent datasets of both watertight and open surfaces. The code and trained models are provided at \url{https://github.com/EnyaHermite/CircNet}.
\end{abstract}
\vspace{-5mm}
\section{Introduction}\label{sec:intro}
\begin{wrapfigure}{r}{0.51
\textwidth}
\includegraphics[width=0.5\textwidth]{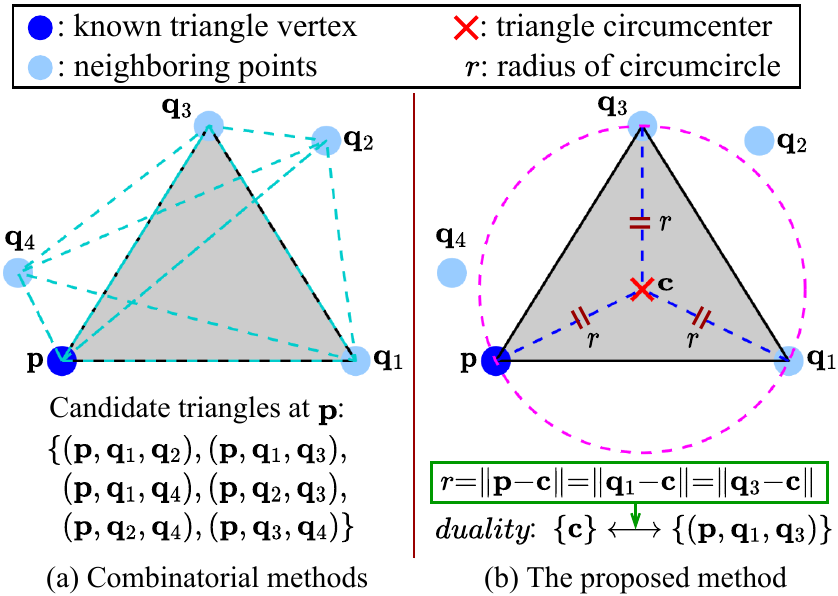}
\vspace{-3mm}
\caption{An example of a point ${\bf p}$ with four neighboring points ${\bf q}_1,{\bf q}_2,{\bf q}_3,{\bf q}_4$. (a) The combinatorial methods propose all of the six triangles incident to ${\bf p}$ as candidate triangles. They are $({\bf p},{\bf q}_1,{\bf q}_2)$,~$({\bf p},{\bf q}_1,{\bf q}_3)$,~$({\bf p},{\bf q}_1,{\bf q}_4)$,~$({\bf p},{\bf q}_2,{\bf q}_3)$,\\~$({\bf p},{\bf q}_2,{\bf q}_4)$,~$({\bf p},{\bf q}_3,{\bf q}_4)$. The neural network has to classify the targeted triangle $({\bf p}, {\bf q}_1,{\bf q}_3)$ out of the six candidates. (b) The proposed method eliminates the candidate proposals by detecting a circumcenter ${\bf c}$ and exploiting its duality with the triangle $({\bf p},{\bf q}_1,{\bf q}_3)$ to identify the targeted triangle.}\label{fig:duality}
\vspace{-3mm}
\end{wrapfigure}

\vspace{-3mm}
Point cloud triangulation~(\citealt{cazals2004delaunay}) aims at reconstructing triangle meshes of object surfaces by adding edge information to their point cloud representations.
The input point clouds are usually produced by either scanning sensors (\textit{e.g.}, LiDAR) or surface sampling methods. 
Compared to implicit surface reconstruction ~(\textit{e.g.}, \citealt{kazhdan2006poisson}), explicit triangulation has the advantages of preserving the original input points and fine-grained details of the surface. Moreover, it does not require oriented normals which are difficult to obtain in practice.  
Recent advances in geometric deep learning have seen widespread applications of neural functions for surface representations (\textit{e.g.},~\citealt{park2019deepsdf, sitzmann2020implicit,sitzmann2020metasdf,erler2020points2surf,gropp2020implicit,atzmon2020sal,atzmon2020sald,BenShabat:CVPR2022,ma2020neural,ma2022surface}). In comparison, only a few methods have been proposed to directly learn triangulation of point clouds by using neural networks. 
This is probably attributed to the combinatorial nature of the triangulation task, hindering the uptake of learning-based methods.  The existing works have to enumerate combinations of candidate triangles around each input point,
and use neural networks to predict their existence in the triangle mesh~(\citealt{sharp2020pointtrinet,liu2020meshing}). 
Figure~\ref{fig:duality}(a) illustrates the local complexity of those combinatorial methods using a point with four neighboring points. Typically, for a point with $K$ neighbors, the combinatorial methods propose ${K \choose 2}$ or $\mathcal{O}(K^2)$ candidate triangles. 

Different from these methods, we propose to exploit
the duality relationship between a triangle and its circumcenter to implement point cloud triangulation. That is, each vertex of a triangle is equally distant to its circumcenter. We use this characteristic to find triangle triplets from their circumcenters. 
Figure.~\ref{fig:duality}(b) shows the duality based on the same example of Fig.~\ref{fig:duality}(a). Our method recovers the vertex triplets of triangle $({\bf p}, {\bf q}_1,{\bf q}_3)$ based on its circumcenter ${\bf c}$ and
the equidistant characteristic, \textit{i.e.}~$\|{\bf p}-{\bf c}\|=\|{\bf q}_1-{\bf c}\|=\|{\bf q}_3-{\bf c}\|$. 
To obtain circumcenters for point cloud triangulation, we introduce a neural network that is able to detect the circumcenters of all triangles in a  mesh. To the best of our knowledge, this is the first single-shot detection architecture for point cloud triangulation. 
We are inspired by the one-stage methods in object detection (\textit{e.g.}, \citealt{liu2016ssd}). 
Unlike previous combinatorial methods, the proposed method removes the requirement for candidate triangles. Specifically, we detect circumcenters in the neighborhood space of each point, under the guidance of a group of anchor priors. The neural network predicts whether a circumcenter exists in the reconstructed mesh, and where it is. We extract triangles induced by the detected circumcenters to obtain a primitive mesh. The final surface mesh is produced by enforcing edge-manifoldness and filling small holes on the primitive mesh.

To validate the proposed method, we train the detection neural network on the ABC dataset (\citealt{koch2019abc}). The trained model is evaluated on the ABC and other datasets, including FAUST~(\citealt{bogo2014faust}), 
MGN~(\citealt{bhatnagar2019multi}), and Matterport3D~(\citealt{Matterport3D}). 
The method not only reconstructs meshes in high quality, but also outperforms the previous learning-based approaches largely in efficiency. It generalizes well to unseen, noisy and non-uniform point clouds.
Our main contributions are summarized below:
\begin{itemize}[leftmargin=*]
\item We introduce the first neural architecture that triangulates point clouds by detecting circumcenters. The duality between a triangle and its circumcenter is exploited afterwards to extract the vertex triplets of each triangle in the mesh. 
\item The proposed neural network is able to reconstruct primitive meshes in milliseconds, due to its single-shot detection pipeline and its removal of candidate proposals. Normals are not required.
\item  The proposed method casts no restriction on the surface topology, \textit{e.g.}, being watertight. Instead, it allows the surfaces to be \textit{open} and have genus (`hole') larger than one. 
\item The method generalizes well to unseen point clouds including those of large-scale scenes. It is robust to non-uniform and noisy data. These indicate its promise for real-world applications. 
\end{itemize}
\vspace{-3mm}
\section{Related Work}
\vspace{-2mm}
Point cloud triangulation and implicit surface functions are two important research directions surface reconstruction. Their major difference is that the former preserves input points, while the latter does not. Alpha shapes~(\citealt{edelsbrunner1994three}) and the ball pivoting algorithm~(\citealt{bernardini1999ball}) are representatives among the traditional methods in point cloud triangulation. The Poisson surface reconstruction~(\citealt{kazhdan2006poisson,kazhdan2013screened}) is a classical approach in implicit surface functions, but it depends on oriented normals of the input points for good performance. Marching Cubes~(\citealt{lorensen1987marching}) and Dual Contouring~(\citealt{ju2002dual}) are restricted to extract the triangle meshes of isosurfaces from their signed distance fields. We refer interested readers to surveys~\citealt{berger2017survey,cheng2013delaunay,newman2006survey} for more in depth discussions of the traditional surface reconstruction methods. 
\vspace{-2mm}
\subsection{Implicit Neural Functions}
\vspace{-2mm}
Implicit neural functions are techniques in geometric deep learning. They exploit the approximation power of neural networks to represent shapes and scenes as level-sets of continuous functions~(\citealt{atzmon2019controlling}). Existing literature trains networks to learn either occupancy functions~(\citealt{mescheder2019occupancy,peng2020convolutional}) or distance functions~(\citealt{park2019deepsdf,sitzmann2020implicit,atzmon2020sal}). 
Since the introduction of DeepSDF~(\citealt{park2019deepsdf}), major advances in this research field include simplifying requirements on the ground-truth (\textit{e.g.}, signs of distances)~(\citealt{atzmon2020sal,atzmon2020sald}), exploring the high-frequency features~(\citealt{sitzmann2020implicit,tancik2020fourier}), and improving loss functions for better surface representation~(\citealt{gropp2020implicit,BenShabat:CVPR2022}).
Implicit functions generate volumetric occupancy or distance fields of isosurfaces, and are followed by Marching Cubes for mesh extraction. Methods in this research direction usually require oriented normals of points, and tend to oversmooth sharp details on the surface. In addition, their neural networks have to be trained with careful initialisation, and are slow at inference stage due to the dense point queries.

There are also methods extending the traditional Marching Cubes, Poisson surface reconstruction, and Dual Contouring using neural networks~(\citealt{liao2018deep, chen2021neural,peng2021shape,chen2022neural}). 
Altogether, they demand interpolations of triangle vertices to reconstruct the mesh, and are unlikely to preserve the fine-grained details carried by the input points.
\vspace{-2mm}
\subsection{Learning-based Point Cloud Triangulation}\label{related:existing_PCT}
\vspace{-2mm}
Compared to their popularity in implicit surface functions, neural networks are much less studied in point cloud triangulation. Existing methods typically reconstruct the triangle mesh by pre-establishing the complete combinations of candidate triangles or parameterizing the local surface patches into the 2D space.
\citet{sharp2020pointtrinet} propose a two-component architecture, PointTriNet. It utilizes a proposal network to recommend candidate triangles, and determines their existence in the triangle mesh with a classification network. 
By comparing the geodesic and Euclidean distances, 
\citet{liu2020meshing} introduce a metric IER to indicate the triangle presence in the reconstructed mesh. They train a neural network to predict the IER values 
of $\mathcal{O}(K^2N)$ candidate triangles. Those candidates are established offline via $K$-nearest neighbor~($K$NN) search~(\citealt{preparata2012computational}). Here $N$ represents the total number of points in the point cloud. 
 As a classical method in computation geometry, Delaunay triangulation (\citealt{mark2008computational}) guarantees point clouds to be triangulated into manifold meshes in the 2D scenario. To exploit this method, \citet{rakotosaona2021learning}
extract $K$NN patches from the 3D point cloud, and parameterize them into 2D space using neural networks. However, dividing point cloud triangulation into learning-based parameterization and computational triangulation makes their method computationally expensive.
Later, \citet{rakotosaona2021differentiable} also study to differentiate the Delaunay triangulation by introducing weighting strategies. Yet, this method is limited to triangular remeshing of manifold surfaces, and not applicable to triangulation of 3D point clouds.

Due to the local computational nature, learning-based triangulation methods generalize well to unseen point clouds of arbitrary shapes and scenes. On the other hand, the implicit neural functions are restricted to shape/surface representations in an instance or category level. It is non-trivial to apply them to cross-category data, \textit{e.g.} mugs to elephants, cars to indoor rooms. We note that even the local implicit functions often generalize poorly~(\citealt{tretschk2020patchnets,chabra2020deep}).
\vspace{-2mm}
\subsection{Learning Triangulation via Circumcenter Detection}
\vspace{-2mm}
Departing significantly from the learning-based triangulation approaches described above, we exploit the duality between a triangle and its circumcenter, and reformulate the combinatorial triangulation as a detection problem of triangle circumcenters. This would facilitate the geomtric deep learning techniques to be applied. Superior to PointTriNet~(\citealt{sharp2020pointtrinet}) which necessitates a two-stage design, our method enables the usage of a one-stage detection pipeline which largely contributes to its efficiency. It has a time complexity of only $\mathcal{O}(tN)$ where $t$ indicates the number of anchors. This is significantly less than the $\mathcal{O}(K^2N)$  of IER~(\citealt{liu2020meshing}).
Similar to existing learning-based methods, we triangulate point clouds based on local $K$NN patches.

\vspace{-2mm}
\section{Method}\label{sec:method}
\vspace{-2mm}
Given a point cloud representation of a surface $\mathcal{P}=\{{\mathbf p}_n\in\mathbb{R}^3\}_{n=1}^{N}$, we focus on triangulating the point cloud into a mesh that reconstructs the underlying surface. 
Unlike implicit surface functions which generate new points for mesh reconstruction~\citep{kazhdan2006poisson}, point cloud triangulation \textit{preserves} the input points by only adding edge information to the existing points. Let $\mathcal{T}=\big\{({\mathbf p}_{n_1},{\mathbf p}_{n_2},{\mathbf p}_{n_3})|{n_1},{n_2},{n_3}\in\{1,2,\dots,N\},{n_1}\neq{n_2}\neq{n_3}\big\}$ be an optimal triangulation of $\mathcal{P}$.  Typically, it reconstructs the surface as an edge-manifold mesh. This indicates that each edge in the triangulation such as $({\mathbf p}_{n_1},{\mathbf p}_{n_2})$ is adjacent to two triangle faces at most. In particular, edges adjacent to one face are the boundary edges. 

\begin{figure}[!t]
  \centering
  \includegraphics[width=0.98\textwidth]{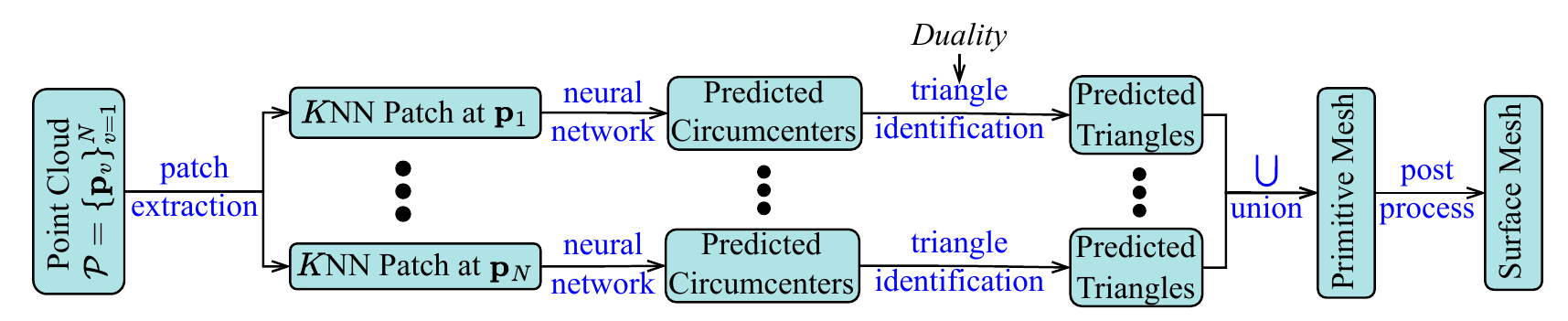}
  \vspace{-3mm}
  \caption{The triangulation process of our method for point cloud $\mathcal{P}$. We extract $K$NN patches to obtain the local geometrics of each point ${\mathbf p}_v$. The neural network detects circumcenters and identifies the adjacent triangles of each ${\mathbf p}_v$ based on their patch inputs. The union of all identified triangles forms the primitive mesh, which is post-processed into an edge-manifold surface mesh.}\label{fig:triangulate_process}
  \vspace{-4.5mm}
\end{figure}
{\bf Overview.} Based on the local geometrics in~\S\ref{subsec:local_geometrics}, we detect circumcenters to predict the \textit{$1$st-order adjacent triangles} of each point. In~\S\ref{subsec:anchor_priors}, we introduce the default anchor priors which help to guide the detection of circumcenters. Details of the detection architecture are presented  in~\S\ref{subsec:network}. We train the neural network with multi-task loss function discussed in~\S\ref{subsec:training}. During inference (\S\ref{subsec:inference}), it triangulates the input point cloud efficiently into a primitive mesh. We post-process the primitive mesh to be an edge-manifold surface mesh. 
Figure~\ref{fig:triangulate_process} summarizes our triangulation process.
\vspace{-2mm}
\subsection{Local Geometrics}\label{subsec:local_geometrics}
\vspace{-2mm}
{\bf $K$NN patch.}
The local geometrics of a point contain rich information for predicting its adjacent triangles. We exploit them as the inputs to our detection network. Specifically, 
we extract 
the local geometrics of each point based on their neighborhoods. 
Let $\mathcal{K}({\mathbf p})=\{{\mathbf q}_{k}|{\mathbf q}_{k}\neq{\mathbf p}\}_{k=1}^{K}$ be a $K$NN patch composed of the $K$-nearest neighbor\footnote{We assume that $K$ is large enough to cover the $1$-ring neighborhood of ${\mathbf p}$ in the triangle mesh.} ($K$NN) points of ${\mathbf p}\in\mathcal{P}$, and $d_0({\mathbf p})>0$ be the distance from ${\mathbf p}$ to its nearest neighbor in
$\mathcal{K}({\mathbf p})$.
To make the neural network robust to density variations in the data, we normalize the $K$NN patch $\mathcal{K}({\mathbf p})$ using a scalar $\eta({\mathbf p})=\frac{\eta_0}{d_0({\mathbf p})}$. Here $\eta_0$ is a hyperparameter controlling the spatial resolution of each $K$NN patch. The normalized patch 
is represented as $\overline{\mathcal{K}}({\mathbf p})=\{\overline{\mathbf q}_{k}|\overline{\mathbf q}_{k}=\eta({\mathbf p})\cdot({\mathbf q}_{k}-{\mathbf p})\}_{i=1}^K$. 
We design graph convolution in \S\ref{subsec:network} to learn global representations of each patch from the input geometrics $\overline{\mathcal{K}}({\mathbf p})$ for circumcenter detection.

{\bf Duality.} Taking $\mathcal{T}$ as the optimal triangulation of $\mathcal{P}$, we denote the adjacent triangles of point ${\mathbf p}$ as $\mathcal{T}({\mathbf p})=\{{\mathbf T}_i({\mathbf p})|{\mathbf p}\in {\mathbf T}_i({\mathbf p})\}$, and the circumcenters of those triangles as $\mathcal{C}({\mathbf p})=\{{\mathbf X}_i({\mathbf p})\}$. 
Our network learns to detect the circumcenters and then extract the adjacent triangles. Let $\widehat{\mathcal{C}}({\mathbf p})=\{\widehat{{\mathbf X}}_m({\mathbf p})\}_{m=1}^M$, $\widehat{\mathcal{T}}({\mathbf p})=\{\widehat{{\mathbf T}}_m({\mathbf p})|{\mathbf p}\in\widehat{{\mathbf T}}_m({\mathbf p})\}_{m=1}^M$ be their respective predictions.  
To extract the triangle triplets $\widehat{{\mathbf T}}_m({\mathbf p})$ based on 
$\widehat{{\mathbf X}}_m({\mathbf p})$, we follow the characteristic that the three vertices of a triangle are equidistant to its circumcenter. 
In practice, the equidistant characteristic has to be applied with 
approximations due to the imperfections of the predicted $\widehat{{\mathbf X}}_m({\mathbf p})$. 
We compute the distance from ${\mathbf p}$ and each of its neighbor point ${\mathbf q}_k\in\mathcal{K}({\mathbf p})$ to a detected circumcenter  $\widehat{{\mathbf X}}_m({\mathbf p})$ as
 \begin{align}
   d_m({\mathbf p}) = \|{\mathbf p}-\widehat{{\mathbf X}}_m({\mathbf p})\|_2,~d_m({\mathbf q}_k) &= \|{\mathbf q}_k-\widehat{{\mathbf X}}_m({\mathbf p})\|_2.
\end{align}
The triangle vertices are determined by the  difference between  distances $d_m({\mathbf q}_k)$ and $d_{m}({\mathbf p})$. 
Let
\begin{align}
 \delta_m({\mathbf q}_k,{\mathbf p})&=\big|d_{m}({\mathbf q}_k)-d_{m}({\mathbf p})\big|,\\
 \Delta_m({\mathbf p}) &=\big\{\delta_m({\mathbf q}_k,{\mathbf p})|{\mathbf q}_k\in\mathcal{K}({\mathbf p})\big\}.
\end{align}
We recover the triangle triplets by selecting the two points ${\mathbf q}_{u},{\mathbf q}_{v}$ which induce the two smallest $\delta_m(\cdot,{\mathbf p})$ in $\Delta_m({\mathbf p})$. Finally, the triangle associated to  $\widehat{{\mathbf X}}_m({\mathbf p})$ is identified as $\widehat{{\mathbf T}}_m({\mathbf p})=({\mathbf p},{\mathbf q}_{u},{\mathbf q}_{v})$. 
\vspace{-2mm}
\subsection{Anchor Priors}\label{subsec:anchor_priors}
\vspace{-2mm}
We use multiple \textit{anchor points} to partition the neighborhood space of each point into different cells. 
The predefined anchor points and cells guide the neural network in its detection of circumcenters. We specify the anchor points in the spherical coordinate system $(\rho,\theta,\phi)$ as it has fixed ranges along the azimuth ($\theta$) and inclination ($\phi$) directions,   \textit{i.e.}~$\theta\in(-\pi,\pi],~\phi\in[-\frac{\pi}{2},\frac{\pi}{2}]$. 
Regarding the radius ($\rho$) direction, we determine its range according to the distributions of circumcenters in the training data, denoted as $(0,R]$. With known ranges, we split the $\rho,\theta,\phi$ each uniformly using fixed steps $\Delta\rho,\Delta\theta,\Delta\phi$, respectively. This results in the number of splits for $\rho, \theta, \phi$ to be $t_\rho=\lceil\frac{R}{\Delta\rho}\rceil$, $t_\theta=\lceil\frac{2\pi}{\Delta\theta}\rceil$, $t_\phi=\lceil\frac{\pi}{\Delta\phi}\rceil$, and a total number of $t=t_\rho\times t_\theta\times t_\phi$ anchor points. 
We represent them as $\mathcal{A}=\big\{{\mathbf a}_j=\big(a_{j_1}^\rho,a_{j_2}^\theta,a_{j_3}^\phi\big)\big\}_{j=1}^t$, where $a_{j_1}^\rho,a_{j_2}^\theta,a_{j_3}^\phi$ each are explicitly defined as 
\begin{align}\label{eq:anchor_points}
\begin{cases}
a_{j_1}^\rho=\frac{\Delta\rho}{2}+(j_1-1)\Delta\rho,~j_1\in\{1,\cdots,t_\rho\},\\[4pt]
a_{j_2}^\theta=\frac{\Delta\theta}{2}+(j_2-1)\Delta\theta,~j_2\in\{1,\cdots,t_\theta\},\\[4pt]
a_{j_3}^\phi= \frac{\Delta\phi}{2}+(j_3-1)\Delta\phi,~j_3\in\{1,\cdots,t_\phi\}.  
\end{cases}
\end{align}
\begin{wrapfigure}{r}{0.25
\textwidth}
\vspace{-11mm}
\includegraphics[width=0.25\textwidth]{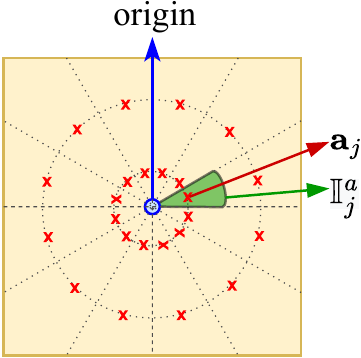}
\vspace{-6mm}
\caption{An example of the anchor priors in 2D. $\Delta\theta=\frac{\pi}{6}$ is used. The anchor points are plotted in red cross, and the anchor cell is colorized as green.}\label{fig:anchor_priors}
\vspace{-14.5mm}
\end{wrapfigure} 
For each anchor point ${\mathbf a}_j=(a_{j_1}^\rho,a_{j_2}^\theta,a_{j_3}^\phi)$, we associate it with an anchor cell defined by the partitioned space 
${\mathbb I}_j^a=I_{j_1}^\rho\times I_{j_2}^\theta\times I_{j_3}^\phi$. 
The anchor cells play an important role in the matching of circumcenters and anchor points (\S\ref{subsec:training}). 
We specify the intervals $I_{j_1}^\rho,I_{j_2}^\theta,I_{j_3}^\phi$ as 
\begin{align}\label{eq:anchor_cells}
\begin{cases}
I_{j_1}^\rho=\big[a^\rho_{j_1}-\frac{\Delta\rho}{2},a^\rho_{j_1}+\frac{\Delta\rho}{2}\big],\\[5pt]
I_{j_2}^\theta=\big[a^\theta_{j_2}-\frac{\Delta\theta}{2},a^\theta_{j_2}+\frac{\Delta\theta}{2}\big],\\[5pt]
I_{j_3}^\phi=\big[a^\phi_{j_3}-\frac{\Delta\phi}{2},a^\phi_{j_3}+\frac{\Delta\phi}{2}\big].
\end{cases}
\end{align}
See Fig.~\ref{fig:anchor_priors} for an example of the anchor points and cells in 2D (\textit{i.e.}~no elevation  direction). With the usage of anchor points and cells, we reduce the combinatorial triangulation of complexity $\mathcal{O}(K^2N)$ to a dual problem of complexity $\mathcal{O}(t N)$. Empirically, $t\ll K^2$. 
Alternative methods for defining the anchor points can be in the Cartesian coordinate system or using the data clustering techniques~(\citealt{bishop2006pattern}).
Yet, they either lead to larger $t$ and hence higher complexity, or make the anchor definition and network training complicated. 
\vspace{-2mm}
\subsection{Network Design}\label{subsec:network}
\begin{figure}[!t]
  \centering
  \includegraphics[width=0.92\textwidth]{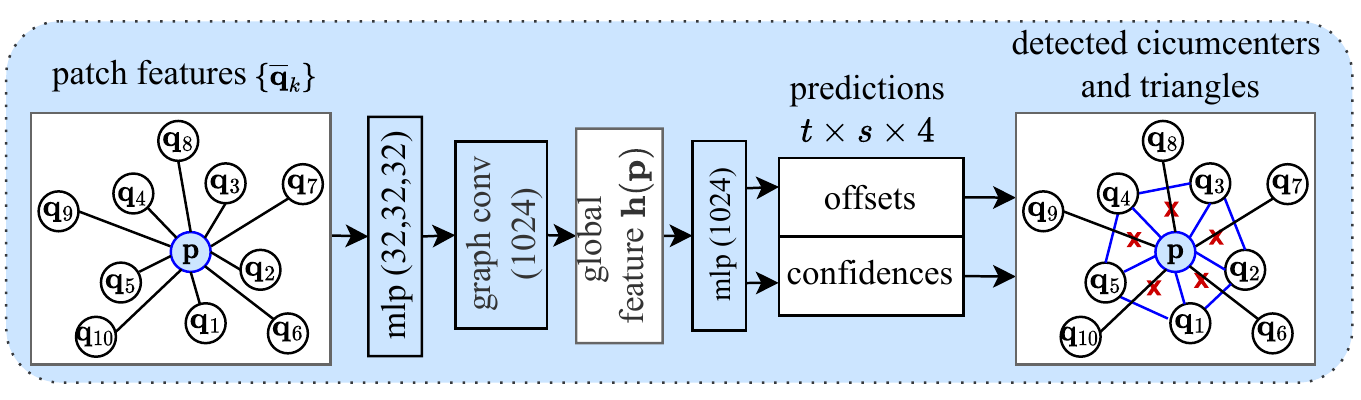}
  \vspace{-3mm}
  \caption{The neural network for circumcenter detection. It learns a global feature representation for each patch using multi-layer perceptrons and the proposed graph convolution. The global feature ${\mathbf h}({\mathbf p})$ is used to make predictions of the circumcenters. We show in the right a toy triangulation of the input patch at ${\bf p}$. The detected circumcenters are in red cross and the triangle edges are in blue.  }\label{fig:anchor_network}
  \vspace{-3mm}
\end{figure}

\vspace{-2mm}
Based on the normalized $K$NN patch $\overline{\mathcal{K}}({\mathbf p})$ of a point ${\mathbf p}$, we design a neural network that is able to predict the circumcenters for adjacent triangle identification. 
The input to our network is a \textit{star graph} which has point ${\mathbf p}$ as its internal node and the neighborhoods $\{{\mathbf q}_k\}$ as its leaf nodes. We present the graph convolution below to encode local geometrics of ${\mathbf p}$ into a global feature representation. The depthwise separable convolution~(\citealt{chollet2017xception}) is explored here. Let $\beta$ be the depth multiplier, ${\mathbf h}^{l-1}({\mathbf q}_k)$ be the input features of point ${\mathbf q}_k$ at layer $l$,  $C_{in}$ be the dimension of ${\mathbf h}^{l-1}({\mathbf q}_k)$, and $C_{out}$ be the expected dimension of output features.
We compute the output features ${\mathbf h}^{l}({\mathbf p})$ of ${\mathbf p}$ as  
\begin{align}\label{eq:convolution}
{\mathbf h}^{l}({\mathbf p}) &= \sum_{k=1}^{K}\sum_{i=1}^{\beta}{\mathbf W}_{i2}\Big(\big({\mathbf W}_{i1}{\mathbf h}^{l-1}({\mathbf q}_k)\big)\odot {\mathbf h}^{l-1}({\mathbf q}_k)\Big) + {\mathbf b}.
\end{align}
Here ${\mathbf W}_{i1},{\mathbf W}_{i2},{\mathbf b}$ 
are learnable parameters in the graph convolution. 
The sizes of ${\mathbf W}_{i1}, {\mathbf W}_{i2}$ are $C_{in}\times C_{in}, C_{out}\times C_{in}$ respectively, 
and the length of ${\mathbf b}$ is $C_{out}$. 
We use the graph convolution only once to calculate global feature representations of each patch. The positional encoding in \citep{mildenhall2021nerf} is also employed to transform the $(x,y,z)$ coordinates into high-frequency input signals. 
Figure~\ref{fig:anchor_network}(b) shows the configurations of our neural detection architecture.
The ultimate output of our network is a tensor of size $t\times s\times4$, where $s$ indicates the number of circumcenters we predict in each anchor cell, and $4$ contains confidence ($z$) about the existence of circumcenter in a cell and its predicted coordinates in 3D. 
Because the usage of one-stage pipeline, the proposed network can detect circumcenters and adjacent triangles efficiently.
\vspace{-3mm}
\subsection{Training}\label{subsec:training}
\vspace{-3mm}
We train the neural network based on multi-task loss functions, similar to those exploited in object detection~(\citealt{liu2016ssd}). The binary cross-entropy is used for classifying the anchor cells, and the smooth L1 loss~(\citealt{girshick2015fast}) is applied for localizing the circumcenters.

{\bf Matching strategy.} 
To train the network accordingly, we have to match the ground-truth circumcenters to the anchor points. Thus, we transform the ground-truth circumcenters $\mathcal{C}({\mathbf p})=\{{\mathbf X}_i({\mathbf p})\}$ into a spherical coordinate system centered at ${\mathbf p}$, denoted as $\mathcal{S}({\mathbf p})=\{(\rho_i,\theta_i,\phi_i)\}$ where $(\rho_i,\theta_i,\phi_i)$ are the spherical coordinates of ${\mathbf X}_i({\mathbf p})-{\mathbf p}$. For each ${\mathbf X}_i({\mathbf p})$, we match it to the anchor point ${\mathbf a}_*=(a^\rho_{*_1},a^\theta_{*_2},a^\phi_{*_3})$ 
if $(\rho_i,\theta_i,\phi_i)\in {\mathbb I}^a_*$.
In this way, it is possible that an anchor point is matched to multiple circumcenters. That is why we allow multiple circumcenters ($s$) to be detected in a single anchor cell. 
The proposed network detects each {\color{black}ground-truth} circumcenter ${\mathbf X}_i({\mathbf p})\in\mathcal{C}({\mathbf p})$ by predicting its parameterized offsets
${\mathbf g}_i({\mathbf p})=(g_i^\rho,g_i^\theta,g_i^\phi)$, defined as
\begin{align}\label{equ::offset}
g_i^\rho = \frac{\rho_i-a^\rho_*}{\Delta\rho},~g_i^\theta = \frac{\theta_i-a^\theta_*}{\Delta\theta},~g_i^\phi = \frac{\phi_i-a^\phi_*}{\Delta\phi}. 
\vspace{-2mm}
\end{align} 
{\color{black}The predicted offsets are $(\widehat{g}_m^\rho,\widehat{g}_m^\theta,\widehat{g}_m^\phi)$.} 
We recover its associated circumcenter prediction $\widehat{{\mathbf X}}_m({\mathbf p})\in\widehat{\mathcal{C}}({\mathbf p})$ with proper coordinate transformation. 
Furthermore, by analyzing the distributions of circumcenters in the training set, we find that two predictions per anchor cell (\textit{i.e.}~$s=2$) reaches a good balance between performance and efficiency. 

{\bf Binary cross-entropy loss.}
For each anchor cell,  whether it contains any ground-truth circumcenters or not is a binary classification task. We predict a confidence $z$ for each of them to indicate the existence of circumcenters. 
It can be seen from \S\ref{subsec:anchor_priors} that we define the anchor points \textit{compactly} in the neighborhood space of a point. In reality, the circumcenters of adjacent triangles for each individual point distribute closely around the surface manifold. Such facts result in the majority of anchor cells to be unoccupied.  They comprise the negative samples $\mathcal{N}_{neg}$ in our classification, while the occupied cells comprise the positive samples $\mathcal{N}_{pos}$. Assume $|\mathcal{N}_{pos}|=N_p$, $|\mathcal{N}_{neg}|=N_n$, and $p$ is the probability correlated to the confidence $z$, we compute the binary cross-entropy as 
\vspace{-2mm}
\begin{align}\label{equ::anchor_loss}
\mathcal{L}_1 &= -\frac{1}{N_p}\sum_{i=1}^{N_p} \log(p_i)-\frac{1}{N_n}\sum_{i=1}^{N_n}\log(1-p_i),~~\text{where}~p_i=\frac{1}{1+\exp(-z_i)}.
\end{align}
We employ hard negative mining for better training~(\citealt{liu2016ssd}).

{\bf Smooth L1 loss.} 
As mentioned above, we predict $s=2$ circumcenters inside each cell. Let $\mathcal{G}_i=\{{\mathbf g}_{i1},\cdots,{\mathbf g}_{i\tau}\}$ be the ground-truth offset coordinates of all circumcenters in the $i$th cell, and $\{\widehat{\mathbf g}_{i1}, \widehat{\mathbf g}_{i2}\}$ be the predicted offsets in the same cell produced by the network. Their computations are consistent with Eq.~(\ref{equ::offset}). If a positive cell contains only one circumcenter (\textit{i.e.}~$\tau=1$), we consider it as having two identical circumcenters inside. If it contains $\tau\geq2$ circumcenters, we match the two predictions to the ground-truth circumcenters that contribute to the minimum loss. 
The localization error is defined as an average of the smooth L1 loss between all pairs of matched ${\mathbf g},\widehat{\mathbf g}$, \textit{i.e.}   
\begin{align}\label{equ::loc_loss}
\mathcal{L}_2 =\frac{1}{N_p}
&\sum_{i=1}^{N_p}\min_{(a,b)\in P(\tau,2)}\big(\text{smooth}_{L1}({\mathbf g}_{ia},\widehat{\mathbf g}_{i1})+\text{smooth}_{L1}({\mathbf g}_{ib},\widehat{\mathbf g}_{i2})\big).
\end{align}
$P(\tau,2)$ represents the permutation of all $2$ selections from $\tau$ elements, \textit{e.g.},~$P(2,2)=\{(1,2),(2,1)\}$. For the special case of $\tau=1$, we define $P(1,2)=\{(1,1)\}$. The $\text{smooth}_{L1}({\mathbf g}_i,\widehat{\mathbf g}_{i})$ is calculated by summing the scalar losses from each dimension of ${\mathbf g}_i$. 

Eventually, the multi-task loss function of our neural network is formulated as 
$\mathcal{L}=\mathcal{L}_1 + \lambda\mathcal{L}_2,$
where $\lambda$ is a hyperparameter for balancing the different loss terms.

\vspace{-2mm}
\subsection{Inference}\label{subsec:inference}
\vspace{-2mm}
We train the network on local patches, but performing the inference on the complete point cloud. After predicting the adjacent triangles of each point, we take their union to form the primitive triangulation (mesh). Since the primitive mesh is produced regardless of topological constraint, we apply post-processing to make it edge-manifold and fill the small holes. For convenience, we provide C 
implementations with Python interface for the post-processing. 

\begin{table}[b]
    \centering
    \vspace{-8mm}
    \caption{\color{black}Surface quality of different triangulation methods on the ABC test set. For each metric, we report the average results across all meshes. The total triangulation time of each learning-based method is reported on the largest point cloud whose size is provided for reference. We also report the network inference time of CircNet in brackets [.].}\label{table:ABC}
    \begin{adjustbox}{width=1\textwidth}
    {
    \Huge
    \begin{tabular}{l|c|c|c|c|c|c|c|c|c}
    \hline
    \multirow{3}{*}{Method}& \multicolumn{7}{c|}{Surface Quality} & \multicolumn{2}{c}{Efficiency} \\
    \cline{2-10}
    & \multicolumn{5}{c|}{overall} & \multicolumn{2}{c|}{sharp} & \multirow{2}{*}{\parbox{2.5cm}{\centering max\\\#points}} & \multirow{2}{*}{\parbox{5.7cm}{\centering total\\runtime/seconds}}\\
    \cline{2-8}
    & CD1($\times10^2$)$\downarrow$  & CD2($\times10^5$)$\downarrow$ & F1$\uparrow$ & NC$\uparrow$ & NR$\downarrow$ & ECD1($\times10^2$)$\downarrow$ & EF1$\uparrow$
    &  & \\
    \hline
     $\alpha$-shapes-3\%&0.448 &2.670 &0.836 &0.943 &7.203 &2.628 &0.616 & \multirow{9}{*}{19669}  & 6.555\\
     $\alpha$-shapes-5\%&0.601 &6.972 &0.802 &0.929 &8.625 &3.767 &0.572 & &6.544\\
     ball-pivot (+$\mathbf{n}$)&0.297 &0.684 &0.939 &0.981 &2.244 &0.782 &0.873 & &\textbf{0.347}\\
    PSR (+$\mathbf{n}$)&0.403 &6.700 & 0.894 &0.971  &6.493  &32.402  &0.095  & &12.571 \\
    PSR (+$\mathbf{n}_{gt}$)&0.400 &6.081  &0.901  &0.972 &6.020 &26.160 &0.108 & &12.529 \\
     \cline{1-8}\cline{10-10}
     DSE&0.285&0.548 &0.949 &0.985 & 1.793 &\textbf{0.538} &\textbf{0.929} &  &59.605 \\
     IER &0.289  &0.580  &0.945 &0.983 &1.949 &0.890 &0.914 & &37.683  \\
     PointTriNet&0.288 &0.790 &0.948 &0.984 &1.931 &0.688 &0.926 & &38.063 \\
     \cline{1-8}\cline{10-10}
     CircNet (Prop.)&\textbf{0.284} &\textbf{0.544} &\textbf{0.950} &\textbf{0.985} &\textbf{1.758} &0.708 &0.924 &   &3.316 [0.996] 
     \\
     \hline
    \end{tabular}
    }
    \end{adjustbox}
    \vspace{-5mm}
    \end{table}
\vspace{-3mm}
\section{Experiment}\label{sec:exp}
\vspace{-2mm}

{\bf ABC dataset.} The ABC dataset~(\citealt{koch2019abc}) is a collection of one million CAD models for deep geometry research. It provides clean synthetic meshes with high-quality triangulations. We use the first five chunks in this dataset to create the training and test sets. Each raw mesh is normalized into a unit sphere and decimated with a voxel grid of $0.01$. This results in a total of $9,026$ meshes, and we apply a train/test split of $25\%/75\%$ to validate effectiveness of the proposed model. The model is trained on the ABC training set. We assess it on the ABC test set as well as other unseen datasets. The implementation details are discussed in the supplementary. 

{\bf Evaluation criteria.} We evaluate the overall surface quality of each reconstructed mesh using Chamfer distances (CD1, CD2), F-Score (F1), normal consistancy (NC), and normal reconstruction error (NR) in degrees. We also evaluate their preservation of sharp details on the surface using Edge Chamfer Distance (ECD1) and Edge F-score (EF1), similar to \citep{chen2022neural}. See the supplementary for computational details about those surface quality metrics. To compare the triangulation efficiency of learning-based methods, we report their inference time on the same machine. The number of points in the point cloud is provided for reference.
\vspace{-2mm}
\subsection{Performance}\label{subsec:performance}
\vspace{-2mm}
{\bf ABC test.} Table~\ref{table:ABC} compares the performance of the proposed method, abbreviated as `CircNet', with those of the traditional triangulation methods, {\color{black}\textit{i.e.,}~$\alpha$-shapes~(\citealt{edelsbrunner1994three}) and ball-pivot~(\citealt{bernardini1999ball}), the implicit surface method PSR~(\citealt{kazhdan2013screened})} and the learning-based triangulation methods~(\textit{i.e.}~PointTriNet~(\citealt{sharp2020pointtrinet}), IER~(\citealt{liu2020meshing}),  DSE~(\citealt{rakotosaona2021learning}). We use the pre-trained weights of other learning-based methods to report their results. It is observed that training those methods from scratch on our ABC training set leads to slightly worse performance. 
We report the performance of $\alpha$-shapes using $\alpha=3\%$ and $\alpha=5\%$. 
We note that ball-pivot requires normals to be estimated first. {\color{black}The performance of PSR is reported by using normals ${\bf n}$ estimated from the point cloud, and normals ${\bf n}_{gt}$ computed from the ground-truth mesh.}
In the efficiency report, we also provide the network inference time of CircNet in brackets~[.].
It can be seen that the proposed CircNet is much faster than the other learning methods. Besides, its reconstructed meshes are in high quality.

\begin{figure}[!t]
\centering
\includegraphics[width=0.98\textwidth]{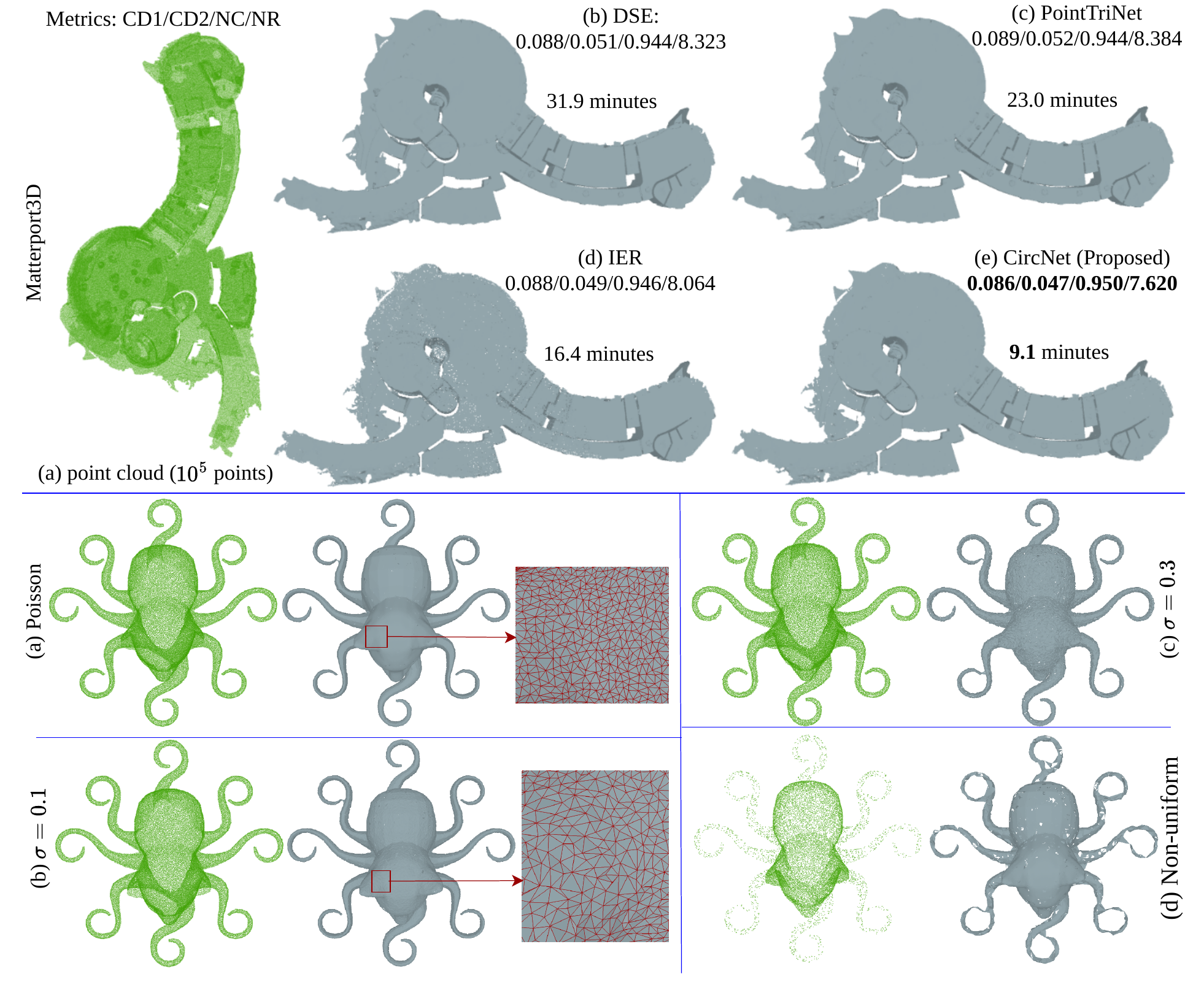}\\
\vspace{-4mm}
\caption{{\color{black}Visualization of the reconstructed meshes. Quantitative results of CD1/CD2/NC/NR and the reconstruction time are reported for different methods over a scene of Matterport3D. 
The proposed CircNet takes the shortest time but reconstructs the scene in best quality. The bottom two rows demonstrate the robustness of CircNet for the fact that it still reconstructs the underlying shape well when the point cloud becomes highly noisy or the non-uniformity becomes severe.}}\label{fig:mesh_visualization}
\vspace{-2mm}
\end{figure}

\begin{table}[t]
\centering
    \caption{\color{black}Method comparison on the watertight meshes of FAUST dataset. The $100$ point clouds in this dataset each have 6890 points. We report the average runtime of each method per sample.}
    \label{table:FAUST}
    \begin{adjustbox}{width=1\textwidth}
    {
     \Huge
    \begin{tabular}{l|c|c|c|c|c|c|c|c|c}
    \hline
    \multirow{3}{*}{Method}& \multicolumn{7}{c|}{Surface Quality} & \multicolumn{2}{c}{Efficiency} \\
    \cline{2-10}
    & \multicolumn{5}{c|}{overall} & \multicolumn{2}{c|}{sharp} & \multirow{2}{*}{\#points} & \multirow{2}{*}{\parbox{5.7cm}{\centering total\\runtime/seconds}}\\
    \cline{2-8}
    & CD1($\times10^2$)$\downarrow$& CD2($\times10^5$)$\downarrow$ & F1$\uparrow$ & NC$\uparrow$ & NR$\downarrow$ &  ECD1($\times10^2$)$\downarrow$ & EF1$\uparrow$
    &  & \\
    \hline
    $\alpha$-shapes-3\%&0.551 &3.689 &0.757 &0.894 &18.197 &7.222 &0.087 & \multirow{10}{*}{6890}  & 0.684\\
    $\alpha$-shapes-5\%&1.225 &19.779 &0.531 &0.807 &27.286 &7.879 &0.056  &  &0.681\\
     ball-pivot (+$\mathbf{n}$)&0.323&1.002 &0.923 &0.970 &6.037 &2.887 &0.184 & &\textbf{0.138}\\
    PSR (+$\mathbf{n}$)&1.119 &39.229  &0.564  &0.863  &21.721  &4.161  &0.438  & &10.674 \\
    PSR (+$\mathbf{n}_{gt}$)&0.427 &4.108  &0.915  &0.969 &10.269 &1.069 &0.810 & &10.643 \\
     \cline{1-8}\cline{10-10}
     DSE&\textbf{0.218}&\textbf{0.307} &\textbf{0.995} &\textbf{0.984} &\textbf{3.910} &\textbf{0.883} &0.801 & &23.792   
     \\
     IER &4.649 &339.565&0.160 &0.786 &31.195 &2.081 &0.376 & &13.949 
     \\
     IER (Poisson)&0.257 &0.406&0.989 &0.973 & 8.692&2.170 &0.456 & & 10.628  
     \\
     PointTriNet&0.219&0.308 &0.995 &0.983 &4.393 &1.233 &0.807  & &13.344 
     \\
      \cline{1-8}\cline{10-10}
     CircNet (Prop.) &0.221 &0.316 &0.993 &0.980 &4.557 &0.939 &\textbf{0.820} && 3.471 [0.382] 
     \\
     \hline
    \end{tabular}
    }
    \end{adjustbox}
    \vspace{-2mm}
    \centering
    \caption{\color{black}Method comparison on the open surfaces of MGN dataset.}
    \label{table:MGN}
    \begin{adjustbox}{width=1\textwidth}
    {
    \Huge
    \begin{tabular}{l|c|c|c|c|c|c|c|c|c}
    \hline
    \multirow{3}{*}{Method}& \multicolumn{7}{c|}{Surface Quality} & \multicolumn{2}{c}{Efficiency} \\
    \cline{2-10}
    & \multicolumn{5}{c|}{overall} & \multicolumn{2}{c|}{sharp} & \multirow{2}{*}{\parbox{2.5cm}{\centering max\\\#points}} & \multirow{2}{*}{\parbox{5.7cm}{\centering total\\runtime/seconds}}\\
    \cline{2-8}
    & CD1($\times10^2$)$\downarrow$& CD2($\times10^5$)$\downarrow$ & F1$\uparrow$ & NC$\uparrow$ & NR$\downarrow$ &  ECD1($\times10^2$)$\downarrow$ & EF1$\uparrow$
    &  & \\
    \hline
    $\alpha$-shapes-3\%&0.517 &2.687 &0.757 &0.927 &14.797 &11.976 &0.046 & \multirow{10}{*}{10116}  & 1.399\\ 
    $\alpha$-shapes-5\%&0.899 &10.322 &0.578 &0.883 &20.233 &14.227 &0.023  &  & 1.390\\
     ball-pivot (+$\mathbf{n}$)&0.462&4.917&0.844&0.974&5.803&11.847&0.083 & & \textbf{0.279}\\
    PSR (+$\mathbf{n}$)&1.319 &18.542  &0.356  &0.892  &18.329  &12.286  &0.071  & &10.389 \\
    PSR (+$\mathbf{n}_{gt}$)&1.077 &10.481  &0.402  &0.948 &12.224 &7.912 &0.137 & &10.623 \\
     \cline{1-8}\cline{10-10}    DSE&0.270&0.530&0.968&\textbf{0.983}&\textbf{3.970}&4.508&0.440& &32.433 
    \\
     IER &0.827&23.464&0.796&0.972&6.220&4.932&0.447& &12.143  
     \\
     IER (Poisson)&0.310&0.635&0.948&0.980&7.073&6.777&0.317& &13.279
     \\
     PointTriNet&0.272&0.562& 0.967&0.981& 4.398&5.936 &0.399 &  &19.906
     \\
      \cline{1-8}\cline{10-10}
     CircNet (Prop.) &\textbf{0.269} &\textbf{0.512}&\textbf{0.968}&0.981&4.230 &\textbf{3.231} &\textbf{0.486} && 4.536 [0.535] 
     \\
     \hline
    \end{tabular}
    }
    \end{adjustbox}
    \vspace{-2mm}
    \centering
    \caption{Our results on Matterport3D, as well as the uniform, non-uniform and noisy data.}
    \label{table:Mat3D_robustABC}
    \begin{adjustbox}{width=1\textwidth}
    {
    \Huge
    \begin{tabular}{l|l|c|c|c|c|c|c|c|c|c}
    \hline
    \multicolumn{2}{l|}{\multirow{3}{*}{Data}}& \multicolumn{7}{c|}{Surface Quality} & \multicolumn{2}{c}{Efficiency} \\
    \cline{3-11}
   \multicolumn{2}{l|}{} & \multicolumn{5}{c|}{overall} & \multicolumn{2}{c|}{sharp} & \multirow{2}{*}{\parbox{2.5cm}{\centering \#points}}& \multirow{2}{*}{\parbox{5.7cm}{\centering total\\runtime/seconds}}\\
    \cline{3-9}
   \multicolumn{2}{l|}{} & CD1($\times10^2$)$\downarrow$& CD2($\times10^5$)$\downarrow$ & F1$\uparrow$ & NC$\uparrow$ & NR$\downarrow$ &  ECD1($\times10^2$)$\downarrow$ & EF1$\uparrow$
    &  & \\
    \hline
     \multicolumn{2}{l|}{Matterport3D} & 0.151 &0.144 &0.999 &0.933 &10.870 &0.249 &0.974 &$5\times10^5$ & $\sim$9.7 minutes   \\
     \hline
     \hline
    \multirow{5}{*}{\rotatebox[origin=c]{90}{\parbox{4cm}{\centering Robustness\\(ABC)}}}&poisson&  0.278 &0.533 &0.949 &0.976 &4.033&1.281&0.660 & $10^4$& 3.190 [0.562] 
    \\
    &$\sigma=0.1$& 0.328 &0.720 &0.908 &0.965 &10.053 &1.500 &0.630 & $10^4$ &3.368 [0.499]
    \\ 
    &$\sigma=0.2$& 0.419 &1.245 &0.793 &0.931 &16.735 &4.396 &0.467 &$10^4$ &5.921 [0.549] 
    \\
    &$\sigma=0.3$& 0.523 &2.076 &0.689 &0.880 &23.287 &6.790 &0.294 &$10^4$ &9.776 [0.627] 
    \\
    &non-uniform& 0.395 & 1.852 &0.858 & 0.942 & 8.693 & 3.490 & 0.457 & 5085 (mean)  & 1.958 [0.525]
    \\
    \hline
    \end{tabular}
    }
    \end{adjustbox}
    \vspace{-10mm}
\end{table}
{\bf Generalization.} We validate generalization of the proposed CircNet using unseen datasets. 
Those include FAUST~(\citealt{bogo2014faust}), a dataset of watertight meshes of human bodies; MGN~(\citealt{bhatnagar2019multi}), a dataset of open meshes of clothes; and several rooms of Matterport3D~(\citealt{Matterport3D}), a dataset of large-scale scenes reconstructed from RGB-D sequences. In all circumstances, the learning-based methods are evaluated without any fine-tuning. We report the results on FAUST and MGN in Table~\ref{table:FAUST} and Table~\ref{table:MGN}, respectively. It can be noticed that CircNet outperforms the other approaches most of time, especially in the reconstruction of the open surfaces from MGN. We observe that IER prefers uniform point clouds as inputs. Due to the non-uniformity of point clouds in FAUST and MGN, its performance drops significantly. As a reference, we provide its results on the uniformly resampled point clouds generated by Poisson-disk sampling~(\citealt{bridson2007fast}). 
For Matterport3D, we report the quantitative results of CircNet over $25$ scenes in  Table~\ref{table:Mat3D_robustABC}.
The input point clouds are generated by uniformly sampling $5\times10^5$ points from the raw meshes.
We compare the reconstructed meshes of CircNet with others on a large building in Fig.~\ref{fig:mesh_visualization}. 

{\bf Robustness.} To test the robustness of CircNet, we generate point clouds that are uniform, noisy and non-uniform using $100$ meshes of the ABC test set. Specifically, we apply the Poisson disk sampling to generate uniform point clouds with $10^4$ points each. We also add different levels of Gaussian noise (\textit{i.e.}, $\sigma=0.1, 0.2, 0.3$) to the data to obtain noisy point clouds. To get non-uniform point clouds, we vary the densities of the uniform data along an axis, similar to DSE~(\citealt{rakotosaona2021learning}). 
The quantitative results of CircNet for each category are provided in Table~\ref{table:Mat3D_robustABC}. We report the quantitative comparisons in the supplementary. Figure~\ref{fig:mesh_visualization} shows the reconstructed meshes of CircNet for point clouds of varying densities or with noise using an interesting shape from the Thingi10K dataset~(\citealt{zhou2016thingi10k}). The point cloud data are generated in similar ways. 

\begin{figure}[!t]
\centering
\includegraphics[width=0.94\textwidth]{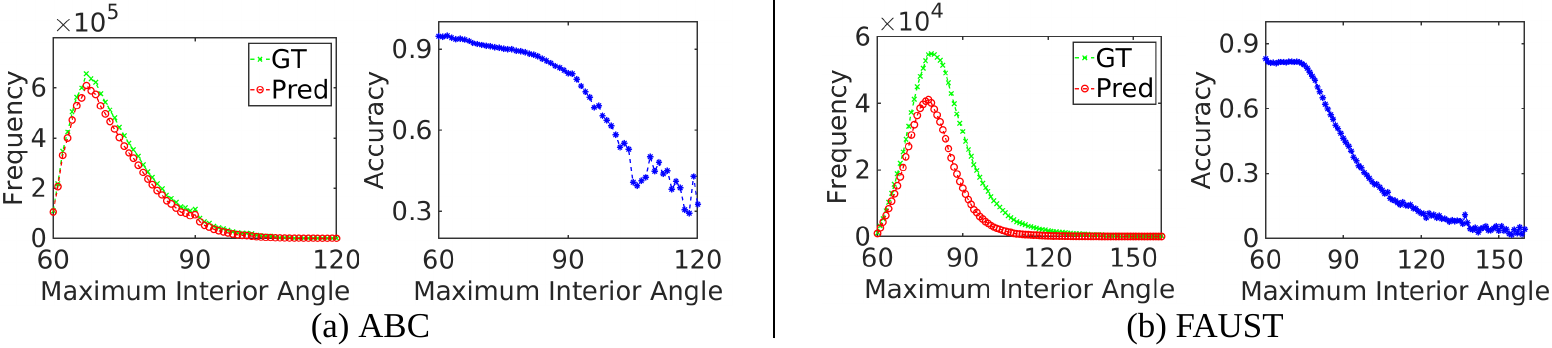}
\vspace{-4.5mm}
\caption{Detection frequency and accuracy with respect to the maximum interior angles (in degree) of ground-truth triangles. We provide results on two datasets, ABC and FAUST. 
For each dataset, the left plot shows the frequencies of both  ground-truth and correctly predicted triangles, while the right plot shows the prediction accuracy. It can be noticed that  triangles in ABC are distributed more closely to equilateral triangles, compared to those in FAUST.
}\label{fig:triAngle_predAcc}
\vspace{-6mm}
\end{figure}

{\bf Limitations.} We analyze detection accuracy of different triangles according to their maximum interior angles. Figure~\ref{fig:triAngle_predAcc} plots such results for the ABC and FAUST datasets. Without loss of generality, we randomly sample $500$ meshes in ABC to plot the results.  We compute the detection accuracy as the ratio between the number of detected triangles and the  ground-truth number. 
When the maximum interior angle becomes large, the radii of the triangle circumcircles can be in an \textit{intractable}~(${\rightarrow}{+}\infty$) range. This makes the circumcenter detection difficult using fixed anchor priors. As expected, the detection accuracy decreases with the increase of the maximum angle. So far, handling edge-manifoldness takes non-trivial amount of time in all learning-based triangulation methods. This is because the classification or detection networks are unaware of any surface topology.

\vspace{-6mm}
\vspace{-2mm}
\section{Conclusion}
\vspace{-4mm}
By exploiting the duality between a triangle and its circumcenter, 
we have introduced a neural detection architecture for point cloud triangulation. 
The proposed network employs a single-shot pipeline, and takes local geometrics as inputs to detect the circumcenters of adjacent triangles of each point.  
We predefine multiple anchor points to guide the detection process. 
Based on the detected circumcenters, we reconstruct the 3D point cloud into a primitive triangle mesh. It  can be simply post-processed into a surface mesh.
Compared to the previous learning-based triangulation methods, the proposed method has lower complexity, single-shot architecture, and does not depend on any traditional method to triangulate the points. We demonstrate that the method, though trained on CAD models, is able to reconstruct unseen point clouds including the large-scale scene data satisfactorily. It is also robust to noisy and non-uniform data. 

\section{Acknowledgement}
This research is funded in part by the ARC Discovery Grant DP220100800 to HL. We thank the anonymous reviewers for their comments on improving the quality of this work.
\bibliography{iclr2023_conference}

\begin{thebibliography}{48}
\providecommand{\natexlab}[1]{#1}
\providecommand{\url}[1]{\texttt{#1}}
\expandafter\ifx\csname urlstyle\endcsname\relax
  \providecommand{\doi}[1]{doi: #1}\else
  \providecommand{\doi}{doi: \begingroup \urlstyle{rm}\Url}\fi

\bibitem[Atzmon \& Lipman(2020{\natexlab{a}})Atzmon and Lipman]{atzmon2020sal}
Matan Atzmon and Yaron Lipman.
\newblock {SAL}: Sign agnostic learning of shapes from raw data.
\newblock In \emph{Proceedings of the IEEE/CVF Conference on Computer Vision
  and Pattern Recognition}, pp.\  2565--2574, 2020{\natexlab{a}}.

\bibitem[Atzmon \& Lipman(2020{\natexlab{b}})Atzmon and Lipman]{atzmon2020sald}
Matan Atzmon and Yaron Lipman.
\newblock {SALD}: Sign agnostic learning with derivatives.
\newblock In \emph{International Conference on Learning Representations},
  2020{\natexlab{b}}.

\bibitem[Atzmon et~al.(2019)Atzmon, Haim, Yariv, Israelov, Maron, and
  Lipman]{atzmon2019controlling}
Matan Atzmon, Niv Haim, Lior Yariv, Ofer Israelov, Haggai Maron, and Yaron
  Lipman.
\newblock Controlling neural level sets.
\newblock \emph{Advances in Neural Information Processing Systems}, 32, 2019.

\bibitem[Ben-Shabat et~al.(2022)Ben-Shabat, Koneputugodage, and
  Gould]{BenShabat:CVPR2022}
Yizhak Ben-Shabat, Chamin~Hewa Koneputugodage, and Stephen Gould.
\newblock {DiGS}: Divergence guided shape implicit neural representation for
  unoriented point clouds.
\newblock In \emph{CVPR}, 2022.

\bibitem[Berger et~al.(2017)Berger, Tagliasacchi, Seversky, Alliez, Guennebaud,
  Levine, Sharf, and Silva]{berger2017survey}
Matthew Berger, Andrea Tagliasacchi, Lee~M Seversky, Pierre Alliez, Gael
  Guennebaud, Joshua~A Levine, Andrei Sharf, and Claudio~T Silva.
\newblock A survey of surface reconstruction from point clouds.
\newblock In \emph{Computer Graphics Forum}, volume~36, pp.\  301--329. Wiley
  Online Library, 2017.

\bibitem[Bernardini et~al.(1999)Bernardini, Mittleman, Rushmeier, Silva, and
  Taubin]{bernardini1999ball}
Fausto Bernardini, Joshua Mittleman, Holly Rushmeier, Cl{\'a}udio Silva, and
  Gabriel Taubin.
\newblock The ball-pivoting algorithm for surface reconstruction.
\newblock \emph{IEEE transactions on visualization and computer graphics},
  5\penalty0 (4):\penalty0 349--359, 1999.

\bibitem[Bhatnagar et~al.(2019)Bhatnagar, Tiwari, Theobalt, and
  Pons-Moll]{bhatnagar2019multi}
Bharat~Lal Bhatnagar, Garvita Tiwari, Christian Theobalt, and Gerard Pons-Moll.
\newblock Multi-garment net: Learning to dress 3d people from images.
\newblock In \emph{proceedings of the IEEE/CVF international conference on
  computer vision}, pp.\  5420--5430, 2019.

\bibitem[Bishop \& Nasrabadi(2006)Bishop and Nasrabadi]{bishop2006pattern}
Christopher~M Bishop and Nasser~M Nasrabadi.
\newblock \emph{Pattern recognition and machine learning}, volume~4.
\newblock Springer, 2006.

\bibitem[Bogo et~al.(2014)Bogo, Romero, Loper, and Black]{bogo2014faust}
Federica Bogo, Javier Romero, Matthew Loper, and Michael~J Black.
\newblock Faust: Dataset and evaluation for 3d mesh registration.
\newblock In \emph{Proceedings of the IEEE conference on computer vision and
  pattern recognition}, pp.\  3794--3801, 2014.

\bibitem[Bridson(2007)]{bridson2007fast}
Robert Bridson.
\newblock Fast poisson disk sampling in arbitrary dimensions.
\newblock \emph{SIGGRAPH sketches}, 10\penalty0 (1):\penalty0 1, 2007.

\bibitem[Cazals \& Giesen(2004)Cazals and Giesen]{cazals2004delaunay}
Fr{\'e}d{\'e}ric Cazals and Joachim Giesen.
\newblock \emph{Delaunay triangulation based surface reconstruction: ideas and
  algorithms}.
\newblock PhD thesis, INRIA, 2004.

\bibitem[Chabra et~al.(2020)Chabra, Lenssen, Ilg, Schmidt, Straub, Lovegrove,
  and Newcombe]{chabra2020deep}
Rohan Chabra, Jan~E Lenssen, Eddy Ilg, Tanner Schmidt, Julian Straub, Steven
  Lovegrove, and Richard Newcombe.
\newblock Deep local shapes: Learning local sdf priors for detailed 3d
  reconstruction.
\newblock In \emph{European Conference on Computer Vision}, pp.\  608--625.
  Springer, 2020.

\bibitem[Chang et~al.(2017)Chang, Dai, Funkhouser, Halber, Niessner, Savva,
  Song, Zeng, and Zhang]{Matterport3D}
Angel Chang, Angela Dai, Thomas Funkhouser, Maciej Halber, Matthias Niessner,
  Manolis Savva, Shuran Song, Andy Zeng, and Yinda Zhang.
\newblock Matterport3d: Learning from rgb-d data in indoor environments.
\newblock \emph{International Conference on 3D Vision (3DV)}, 2017.

\bibitem[Chen \& Zhang(2021)Chen and Zhang]{chen2021neural}
Zhiqin Chen and Hao Zhang.
\newblock Neural marching cubes.
\newblock \emph{ACM Transactions on Graphics (TOG)}, 40\penalty0 (6):\penalty0
  1--15, 2021.

\bibitem[Chen et~al.(2022)Chen, Tagliasacchi, Funkhouser, and
  Zhang]{chen2022neural}
Zhiqin Chen, Andrea Tagliasacchi, Thomas Funkhouser, and Hao Zhang.
\newblock Neural dual contouring.
\newblock \emph{arXiv preprint arXiv:2202.01999}, 2022.

\bibitem[Cheng et~al.(2013)Cheng, Dey, Shewchuk, and Sahni]{cheng2013delaunay}
Siu-Wing Cheng, Tamal~Krishna Dey, Jonathan Shewchuk, and Sartaj Sahni.
\newblock \emph{Delaunay mesh generation}.
\newblock CRC Press Boca Raton, 2013.

\bibitem[Chollet(2017)]{chollet2017xception}
Fran{\c{c}}ois Chollet.
\newblock Xception: Deep learning with depthwise separable convolutions.
\newblock In \emph{Proceedings of the IEEE conference on computer vision and
  pattern recognition}, pp.\  1251--1258, 2017.

\bibitem[Edelsbrunner \& M{\"u}cke(1994)Edelsbrunner and
  M{\"u}cke]{edelsbrunner1994three}
Herbert Edelsbrunner and Ernst~P M{\"u}cke.
\newblock Three-dimensional alpha shapes.
\newblock \emph{ACM Transactions on Graphics (TOG)}, 13\penalty0 (1):\penalty0
  43--72, 1994.

\bibitem[Erler et~al.(2020)Erler, Guerrero, Ohrhallinger, Mitra, and
  Wimmer]{erler2020points2surf}
Philipp Erler, Paul Guerrero, Stefan Ohrhallinger, Niloy~J Mitra, and Michael
  Wimmer.
\newblock Points2surf learning implicit surfaces from point clouds.
\newblock In \emph{European Conference on Computer Vision}, pp.\  108--124.
  Springer, 2020.

\bibitem[Girshick(2015)]{girshick2015fast}
Ross Girshick.
\newblock Fast r-cnn.
\newblock In \emph{Proceedings of the IEEE international conference on computer
  vision}, pp.\  1440--1448, 2015.

\bibitem[Gropp et~al.(2020)Gropp, Yariv, Haim, Atzmon, and
  Lipman]{gropp2020implicit}
Amos Gropp, Lior Yariv, Niv Haim, Matan Atzmon, and Yaron Lipman.
\newblock Implicit geometric regularization for learning shapes.
\newblock In \emph{International Conference on Machine Learning}, pp.\
  3789--3799. PMLR, 2020.

\bibitem[Ju et~al.(2002)Ju, Losasso, Schaefer, and Warren]{ju2002dual}
Tao Ju, Frank Losasso, Scott Schaefer, and Joe Warren.
\newblock Dual contouring of hermite data.
\newblock In \emph{Proceedings of the 29th annual conference on Computer
  graphics and interactive techniques}, pp.\  339--346, 2002.

\bibitem[Kazhdan \& Hoppe(2013)Kazhdan and Hoppe]{kazhdan2013screened}
Michael Kazhdan and Hugues Hoppe.
\newblock Screened poisson surface reconstruction.
\newblock \emph{ACM Transactions on Graphics (ToG)}, 32\penalty0 (3):\penalty0
  1--13, 2013.

\bibitem[Kazhdan et~al.(2006)Kazhdan, Bolitho, and Hoppe]{kazhdan2006poisson}
Michael Kazhdan, Matthew Bolitho, and Hugues Hoppe.
\newblock Poisson surface reconstruction.
\newblock In \emph{Proceedings of the fourth Eurographics symposium on Geometry
  processing}, volume~7, 2006.

\bibitem[Kingma \& Ba(2015)Kingma and Ba]{kingma2015adam}
Diederik~P Kingma and Jimmy Ba.
\newblock Adam: A method for stochastic optimization.
\newblock \emph{International Conference on Learning Representations}, 2015.

\bibitem[Koch et~al.(2019)Koch, Matveev, Jiang, Williams, Artemov, Burnaev,
  Alexa, Zorin, and Panozzo]{koch2019abc}
Sebastian Koch, Albert Matveev, Zhongshi Jiang, Francis Williams, Alexey
  Artemov, Evgeny Burnaev, Marc Alexa, Denis Zorin, and Daniele Panozzo.
\newblock {ABC}: A big {CAD} model dataset for geometric deep learning.
\newblock In \emph{Proceedings of the IEEE/CVF Conference on Computer Vision
  and Pattern Recognition}, pp.\  9601--9611, 2019.

\bibitem[Liao et~al.(2018)Liao, Donne, and Geiger]{liao2018deep}
Yiyi Liao, Simon Donne, and Andreas Geiger.
\newblock Deep marching cubes: Learning explicit surface representations.
\newblock In \emph{Proceedings of the IEEE Conference on Computer Vision and
  Pattern Recognition}, pp.\  2916--2925, 2018.

\bibitem[Liu et~al.(2020)Liu, Zhang, and Su]{liu2020meshing}
Minghua Liu, Xiaoshuai Zhang, and Hao Su.
\newblock Meshing point clouds with predicted intrinsic-extrinsic ratio
  guidance.
\newblock In \emph{European Conference on Computer Vision}, pp.\  68--84.
  Springer, 2020.

\bibitem[Liu et~al.(2016)Liu, Anguelov, Erhan, Szegedy, Reed, Fu, and
  Berg]{liu2016ssd}
Wei Liu, Dragomir Anguelov, Dumitru Erhan, Christian Szegedy, Scott Reed,
  Cheng-Yang Fu, and Alexander~C Berg.
\newblock {SSD}: Single shot multibox detector.
\newblock In \emph{European conference on computer vision}, pp.\  21--37.
  Springer, 2016.

\bibitem[Lorensen \& Cline(1987)Lorensen and Cline]{lorensen1987marching}
William~E Lorensen and Harvey~E Cline.
\newblock Marching cubes: A high resolution 3d surface construction algorithm.
\newblock \emph{ACM siggraph computer graphics}, 21\penalty0 (4):\penalty0
  163--169, 1987.

\bibitem[Ma et~al.(2021)Ma, Han, Liu, and Zwicker]{ma2020neural}
Baorui Ma, Zhizhong Han, Yu-Shen Liu, and Matthias Zwicker.
\newblock Neural-pull: Learning signed distance functions from point clouds by
  learning to pull space onto surfaces.
\newblock \emph{International Conference on Machine Learning}, 2021.

\bibitem[Ma et~al.(2022)Ma, Liu, Zwicker, and Han]{ma2022surface}
Baorui Ma, Yu-Shen Liu, Matthias Zwicker, and Zhizhong Han.
\newblock Surface reconstruction from point clouds by learning predictive
  context priors.
\newblock In \emph{Proceedings of the IEEE/CVF Conference on Computer Vision
  and Pattern Recognition}, pp.\  6326--6337, 2022.

\bibitem[Mark et~al.(2008)Mark, Otfried, Marc, and Mark]{mark2008computational}
de~Berg Mark, Cheong Otfried, van~Kreveld Marc, and Overmars Mark.
\newblock \emph{Computational geometry algorithms and applications}.
\newblock Spinger, 2008.

\bibitem[Mescheder et~al.(2019)Mescheder, Oechsle, Niemeyer, Nowozin, and
  Geiger]{mescheder2019occupancy}
Lars Mescheder, Michael Oechsle, Michael Niemeyer, Sebastian Nowozin, and
  Andreas Geiger.
\newblock Occupancy networks: Learning 3d reconstruction in function space.
\newblock In \emph{Proceedings of the IEEE/CVF conference on computer vision
  and pattern recognition}, pp.\  4460--4470, 2019.

\bibitem[Mildenhall et~al.(2020)Mildenhall, Srinivasan, Tancik, Barron,
  Ramamoorthi, and Ng]{mildenhall2021nerf}
Ben Mildenhall, Pratul~P Srinivasan, Matthew Tancik, Jonathan~T Barron, Ravi
  Ramamoorthi, and Ren Ng.
\newblock Nerf: Representing scenes as neural radiance fields for view
  synthesis.
\newblock In \emph{European Conference on Computer Vision}. Springer, 2020.

\bibitem[Newman \& Yi(2006)Newman and Yi]{newman2006survey}
Timothy~S Newman and Hong Yi.
\newblock A survey of the marching cubes algorithm.
\newblock \emph{Computers \& Graphics}, 30\penalty0 (5):\penalty0 854--879,
  2006.

\bibitem[Park et~al.(2019)Park, Florence, Straub, Newcombe, and
  Lovegrove]{park2019deepsdf}
Jeong~Joon Park, Peter Florence, Julian Straub, Richard Newcombe, and Steven
  Lovegrove.
\newblock Deepsdf: Learning continuous signed distance functions for shape
  representation.
\newblock In \emph{Proceedings of the IEEE/CVF Conference on Computer Vision
  and Pattern Recognition}, pp.\  165--174, 2019.

\bibitem[Peng et~al.(2020)Peng, Niemeyer, Mescheder, Pollefeys, and
  Geiger]{peng2020convolutional}
Songyou Peng, Michael Niemeyer, Lars Mescheder, Marc Pollefeys, and Andreas
  Geiger.
\newblock Convolutional occupancy networks.
\newblock In \emph{European Conference on Computer Vision}, pp.\  523--540.
  Springer, 2020.

\bibitem[Peng et~al.(2021)Peng, Jiang, Liao, Niemeyer, Pollefeys, and
  Geiger]{peng2021shape}
Songyou Peng, Chiyu Jiang, Yiyi Liao, Michael Niemeyer, Marc Pollefeys, and
  Andreas Geiger.
\newblock Shape as points: A differentiable poisson solver.
\newblock \emph{Advances in Neural Information Processing Systems},
  34:\penalty0 13032--13044, 2021.

\bibitem[Preparata \& Shamos(2012)Preparata and
  Shamos]{preparata2012computational}
Franco~P Preparata and Michael~I Shamos.
\newblock \emph{Computational geometry: an introduction}.
\newblock Springer Science \& Business Media, 2012.

\bibitem[Rakotosaona et~al.(2021{\natexlab{a}})Rakotosaona, Aigerman, Mitra,
  Ovsjanikov, and Guerrero]{rakotosaona2021differentiable}
Marie-Julie Rakotosaona, Noam Aigerman, Niloy~J Mitra, Maks Ovsjanikov, and
  Paul Guerrero.
\newblock Differentiable surface triangulation.
\newblock \emph{ACM Transactions on Graphics (TOG)}, 40\penalty0 (6):\penalty0
  1--13, 2021{\natexlab{a}}.

\bibitem[Rakotosaona et~al.(2021{\natexlab{b}})Rakotosaona, Guerrero, Aigerman,
  Mitra, and Ovsjanikov]{rakotosaona2021learning}
Marie-Julie Rakotosaona, Paul Guerrero, Noam Aigerman, Niloy~J Mitra, and Maks
  Ovsjanikov.
\newblock Learning delaunay surface elements for mesh reconstruction.
\newblock In \emph{Proceedings of the IEEE/CVF Conference on Computer Vision
  and Pattern Recognition}, pp.\  22--31, 2021{\natexlab{b}}.

\bibitem[Sharp \& Ovsjanikov(2020)Sharp and Ovsjanikov]{sharp2020pointtrinet}
Nicholas Sharp and Maks Ovsjanikov.
\newblock Pointtrinet: Learned triangulation of 3d point sets.
\newblock In \emph{European Conference on Computer Vision}, pp.\  762--778.
  Springer, 2020.

\bibitem[Sitzmann et~al.(2020{\natexlab{a}})Sitzmann, Chan, Tucker, Snavely,
  and Wetzstein]{sitzmann2020metasdf}
Vincent Sitzmann, Eric Chan, Richard Tucker, Noah Snavely, and Gordon
  Wetzstein.
\newblock Metasdf: Meta-learning signed distance functions.
\newblock \emph{Advances in Neural Information Processing Systems},
  33:\penalty0 10136--10147, 2020{\natexlab{a}}.

\bibitem[Sitzmann et~al.(2020{\natexlab{b}})Sitzmann, Martel, Bergman, Lindell,
  and Wetzstein]{sitzmann2020implicit}
Vincent Sitzmann, Julien Martel, Alexander Bergman, David Lindell, and Gordon
  Wetzstein.
\newblock Implicit neural representations with periodic activation functions.
\newblock \emph{Advances in Neural Information Processing Systems},
  33:\penalty0 7462--7473, 2020{\natexlab{b}}.

\bibitem[Tancik et~al.(2020)Tancik, Srinivasan, Mildenhall, Fridovich-Keil,
  Raghavan, Singhal, Ramamoorthi, Barron, and Ng]{tancik2020fourier}
Matthew Tancik, Pratul Srinivasan, Ben Mildenhall, Sara Fridovich-Keil, Nithin
  Raghavan, Utkarsh Singhal, Ravi Ramamoorthi, Jonathan Barron, and Ren Ng.
\newblock Fourier features let networks learn high frequency functions in low
  dimensional domains.
\newblock \emph{Advances in Neural Information Processing Systems},
  33:\penalty0 7537--7547, 2020.

\bibitem[Tretschk et~al.(2020)Tretschk, Tewari, Golyanik, Zollh{\"o}fer, Stoll,
  and Theobalt]{tretschk2020patchnets}
Edgar Tretschk, Ayush Tewari, Vladislav Golyanik, Michael Zollh{\"o}fer,
  Carsten Stoll, and Christian Theobalt.
\newblock Patchnets: Patch-based generalizable deep implicit 3d shape
  representations.
\newblock In \emph{European Conference on Computer Vision}, pp.\  293--309.
  Springer, 2020.

\bibitem[Zhou \& Jacobson(2016)Zhou and Jacobson]{zhou2016thingi10k}
Qingnan Zhou and Alec Jacobson.
\newblock Thingi10k: A dataset of 10,000 3d-printing models.
\newblock \emph{arXiv preprint arXiv:1605.04797}, 2016.

\end{thebibliography}
\bibliographystyle{iclr2023_conference}

\newpage
\setcounter{table}{0}
\renewcommand{\thetable}{\Alph{table}}
\setcounter{figure}{0}
\renewcommand{\thefigure}{\Alph{figure}}
\section*{\LARGE Appendices}
\begin{figure}[b]
    \centering
    \includegraphics[width=1\textwidth]{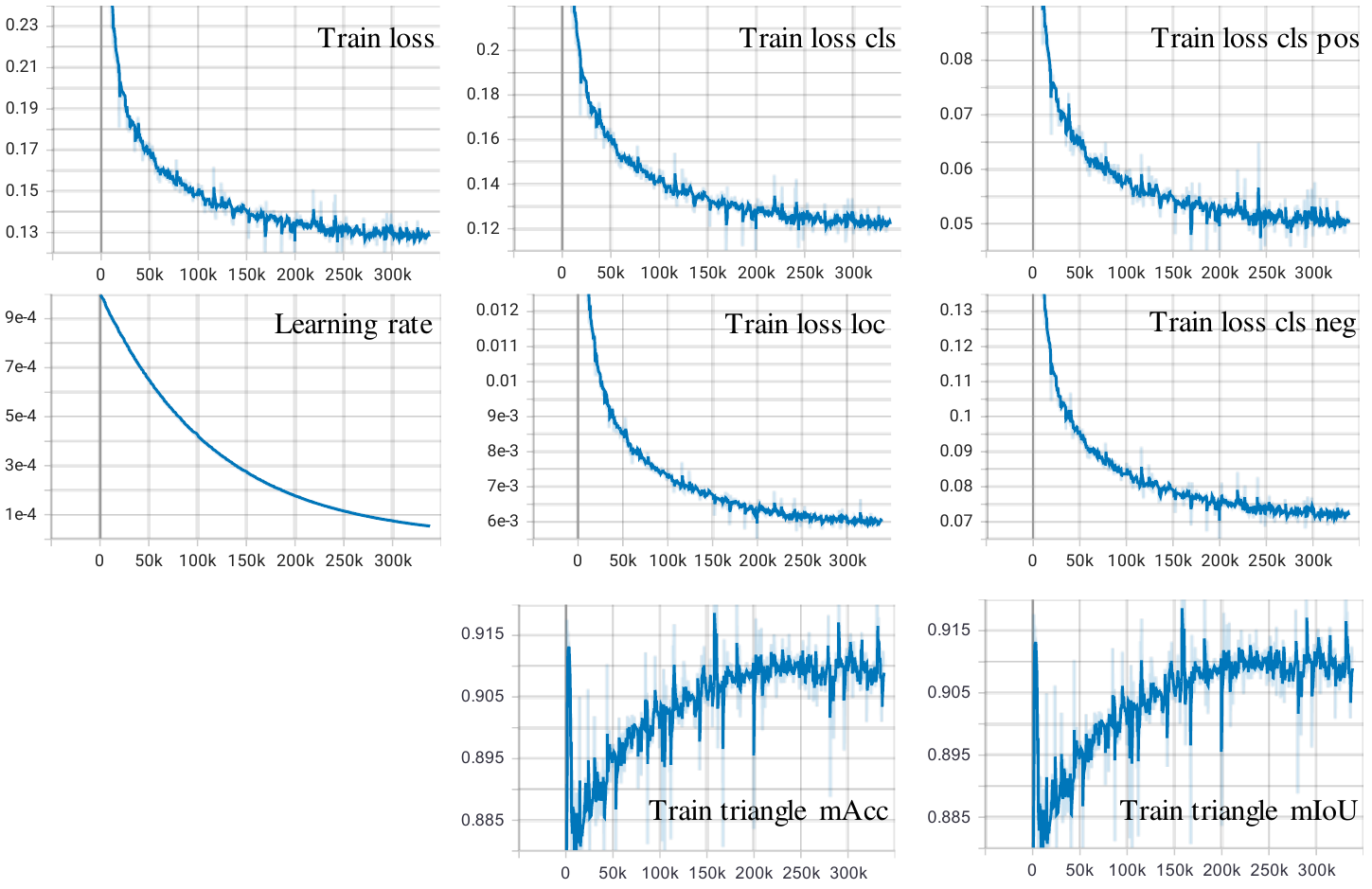}
    \includegraphics[width=1\textwidth]{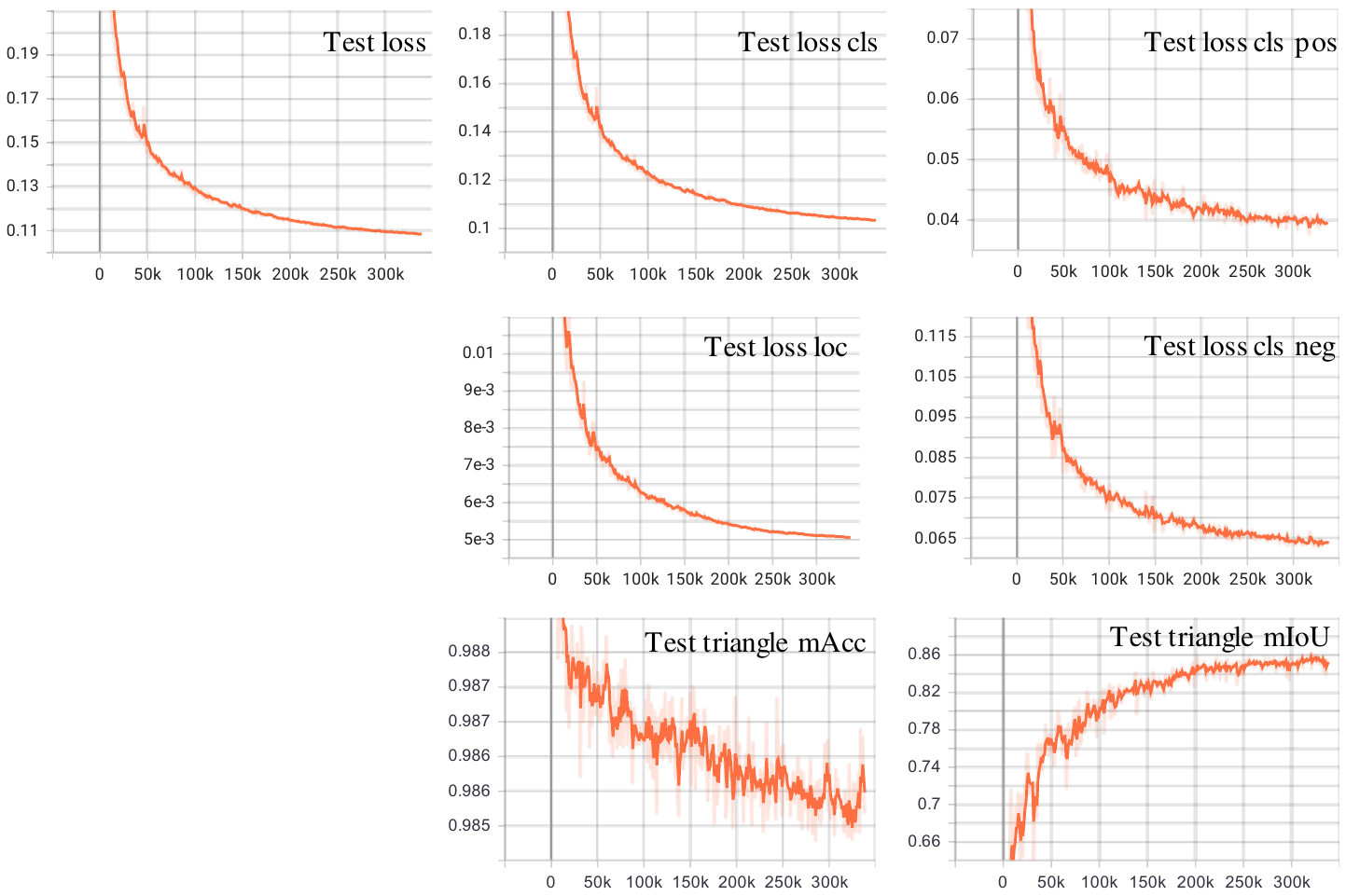}
    \caption{\color{black}Training and test curves. The mAcc and mIoU metrics are computed based on per triangle.}
    \label{fig:training_curve}
\end{figure}

{\bf Implementation details.} We use $\eta_0=0.01$ to adjust the $K$NN patch resolution. To obtain the anchor priors, we set the maximum radius as $R=0.2$~(i.e. $20\times\eta_0$), and the splitting steps along different axes as  $\Delta\rho=0.02$~(i.e. $2\times\eta_0$), $\Delta\theta=\frac{\pi}{6}, \Delta\phi=\frac{\pi}{6}$. It results in a total number of $720$ anchor points. We apply a multiplier of $\beta=16$ for the  graph convolution,  a level of $L=16$ for the positional encoding, and a ratio of 1:20 between positive and negative samples in Eq.~(8). We set $\lambda=1$ in the final loss computation. To train the network, we sample $K$NN patches with $K=50$ randomly from the point cloud. Standard geometric transformations such as scaling, rotating, noisy perturbation, are applied to augment the point cloud. 
We use Adam optimizer~(\citealt{kingma2015adam}) with an initial learning rate of $0.001$, which is decayed by $0.5$ per $8\times10^4$ iterations. Our batch size for the $K$NN patches is $400$.
The number of training parameters in the proposed CircNet is 7.5 million (7.5M). See Fig.~\ref{fig:training_curve} for the training details of our detection network. {\color{black}To demonstrate the quality of our local predictions/triangulations, we report additional metrics in terms of the average accuracy (mAcc) and average intersection-over-union (mIoU) of the $K$NN patches. Per triangle forms a prediction element in those computations.}

{\bf Surface quality.} 
We evaluate the overall quality of each reconstructed mesh using Chamfer Distance~(CD1), squared Chamfer~(CD2) and F-Score (F1), and the quality of their surface normals with Normal Consistency (NC) and Normal Reconstruction error in degrees (NR). 
Preserving sharp edges is also important for the surface reconstruction. We therefore evaluate the Edge Chamfer Distance (ECD1) and Edge F-score (EF1) of the reconstructed meshes. 
Note that the computations of metrics CD1, CD2 and F1 require points to be sampled densely on the entire surface, while those of ECD1 and EF1 utilizes only points sampled near the edges and corners. We follow the convention~(\citealt{sharp2020pointtrinet}) to sample $10^5$ points on the ground-truth and reconstructed meshes for evaluating the quality of shapes. While for the evaluation of large-scale scenes in Matterport3D, one million points are sampled for better coverage of the surface. 

{\color{black}For explicity, we provide the definitions of Chamfer distance and F-score below. Given a ground-truth mesh $\mathcal{T}$ and its reconstructed mesh $\widehat{\mathcal{T}}$, we sample the same number of points (\eg~$10^5$) uniformly on each of them. Let the resulted point clouds be $\mathcal{Q}$ and $\widehat{\mathcal{Q}}$, respectively. The chamfer distance (CD1) is calculated as 
\begin{align}\label{equ::chamfer_distance}
\text{CD1}(\mathcal{Q}, \widehat{\mathcal{Q}}) = \frac{1}{|\mathcal{Q}|}\sum_{\mathbf{x}\in \mathcal{Q}}\min_{\mathbf{y} \in \widehat{\mathcal{Q}}}\|\mathbf{x}-\mathbf{y}\|_2 + \frac{1}{|\widehat{\mathcal{Q}}|}\sum_{\mathbf{x}\in \widehat{\mathcal{Q}}}\min_{\mathbf{y} \in \mathcal{Q}}\|\mathbf{x}-\mathbf{y}\|_2.
\end{align}
The first term measures completeness of the reconstructed mesh, while the second term measures its accuracy.
For the squared chamfer distance (CD2), $\|\mathbf{x}-\mathbf{y}\|_2$ is replaced as $\|\mathbf{x}-\mathbf{y}\|_2^2$. The computation of F-score follows its standard definition, \ie
\begin{align}\label{equ::F_score}
\mathrm{F1} = \frac{2\times\mathrm{recall}\times\mathrm{precision}}{\mathrm{recall} + \mathrm{precision}}.
\end{align}
The recall and precision are calculated as 
\begin{align}\label{equ::F_score}
\mathrm{recall} &= \frac{1}{|\mathcal{Q}|}\sum_{\mathbf{x}\in \mathcal{Q}}{\mathbf 1}\big((\min_{\mathbf{y} \in \widehat{\mathcal{Q}}}\|\mathbf{x}-\mathbf{y}\|_2)<\epsilon\big),\\
\mathrm{precision} &= \frac{1}{|\widehat{\mathcal{Q}}|}\sum_{\mathbf{x}\in \widehat{\mathcal{Q}}}{\mathbf 1}\big((\min_{\mathbf{y} \in \mathcal{Q}}\|\mathbf{x}-\mathbf{y}\|_2)<\epsilon),
\end{align}
where $\epsilon$ is small threshold for the distance and ${\mathbf 1}(\cdot)$ is the indicator function.}

{\bf Efficiency.} In addition to surface quality, a triangulation method is expected to reconstruct high-quality mesh in short time. For efficiency comparison, we report the total reconstruction time of each method on the same machine with one NVIDIA GeForce RTX 2080Ti GPU and AMD Ryzen Threadripper 2990WX CPU. 
The results of alpha-shapes and ball-pivot are computed based on the latest `pymeshlab' (i.e.~2022.2.post2) for Python. To be more specific, suppose the bounding box diagonal of a point cloud is $l_D$. We report the performance of $\alpha$-shapes using $\alpha=3\%\times l_D$ and $\alpha=5\%\times l_D$. For ball-pivoting, the radius is set to be $1\%\times l_D$. {\color{black}The results of PSR are produced using `open3d' (i.e.~0.15.2+c074f5d) by setting depth to $9$.} 

{\bf More results on robustness.}
We show the quantitative comparisons of different learning-based triangulation methods on the ABC point clouds that are  uniform, different levels of noise, and non-uniform distributions in Table~\ref{table:uniform}, Table~\ref{table:noise_sigma_01}, Table~\ref{table:noise_sigma_02}, Table~\ref{table:noise_sigma_03}, Table~\ref{table:non_uniform}, respectively. We also report their performance on data that are sampled under random uniformity in Table~\ref{table:random}. 

\begin{figure}[t]
    \centering
    \includegraphics[width=1\textwidth]{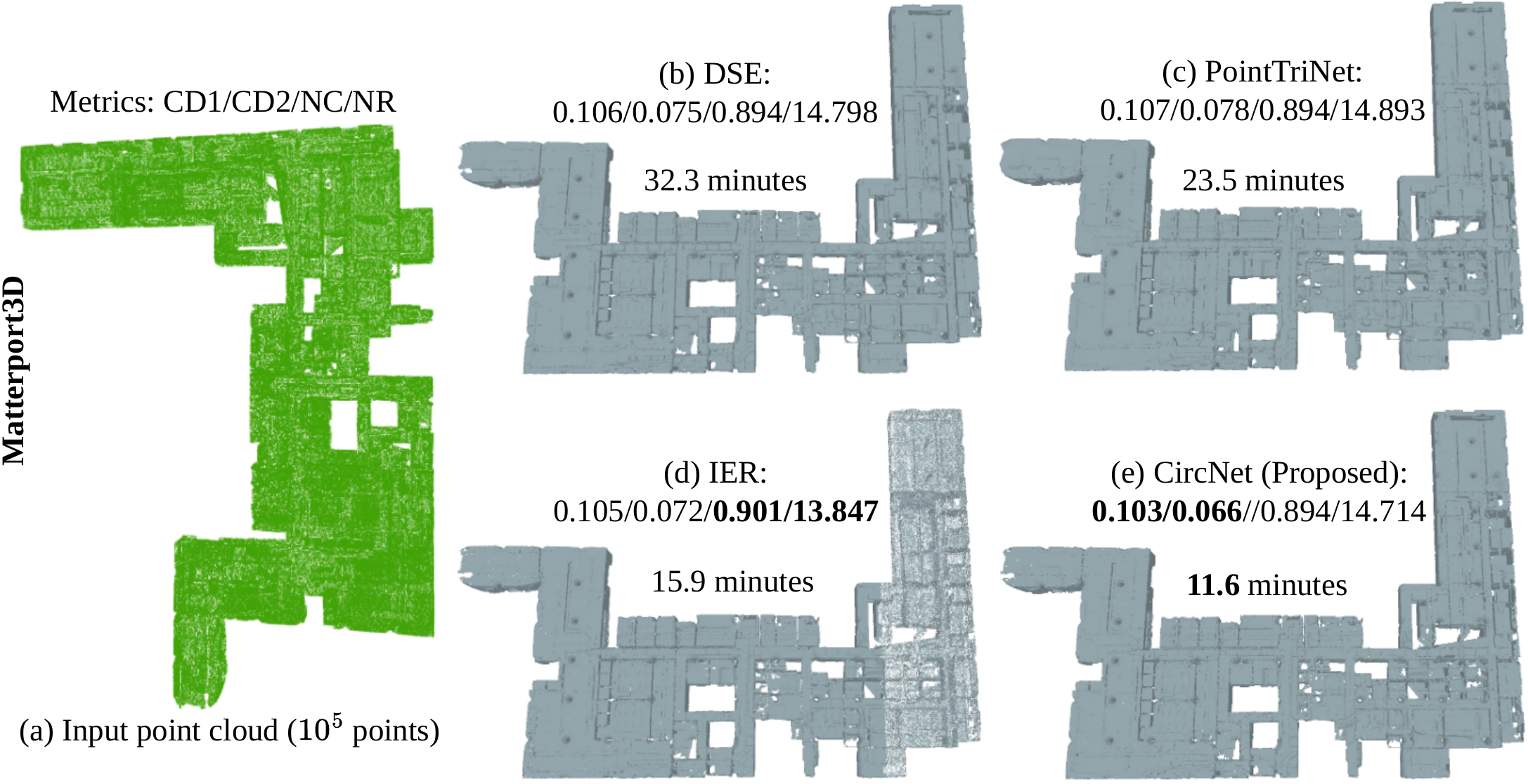}
    \caption{Reconstructed scene meshes of different learning-based triangulation methods. We compare their surface quality using the metrics CD1, CD2, NC, and NR. It can be seen that the mesh of CircNet has the lowest Chamfer distances.}
    \label{fig:Mat3D_scene2}
\end{figure}

\begin{figure}[b]
    \centering    \includegraphics[width=1\textwidth]{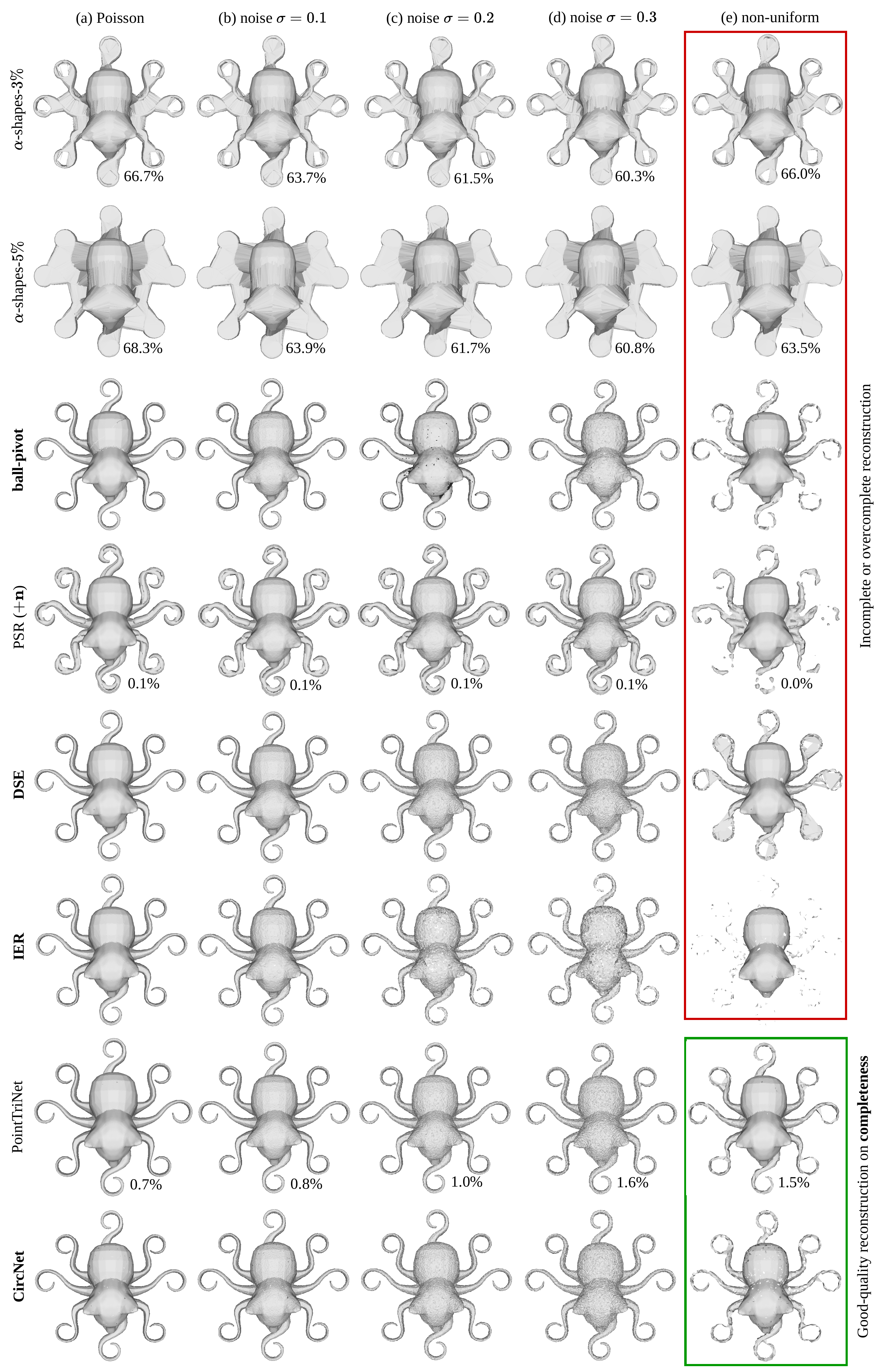}
    \vspace{-8mm}
    \caption{\color{black}Reconstructed surface shapes of different methods for noisy and non-uniform point clouds. Among those methods, ball-pivot, DSE, IER and the proposed CircNet guarantee an edge-manifold surface as output, while the other methods produce non-manifold meshes. We show the percentage of non-manifold edges of each reconstructed mesh right below it.}
    \label{fig:more_vis_shape1}
\end{figure}

\begin{figure}[b]
    \centering    \includegraphics[width=0.88\textwidth]{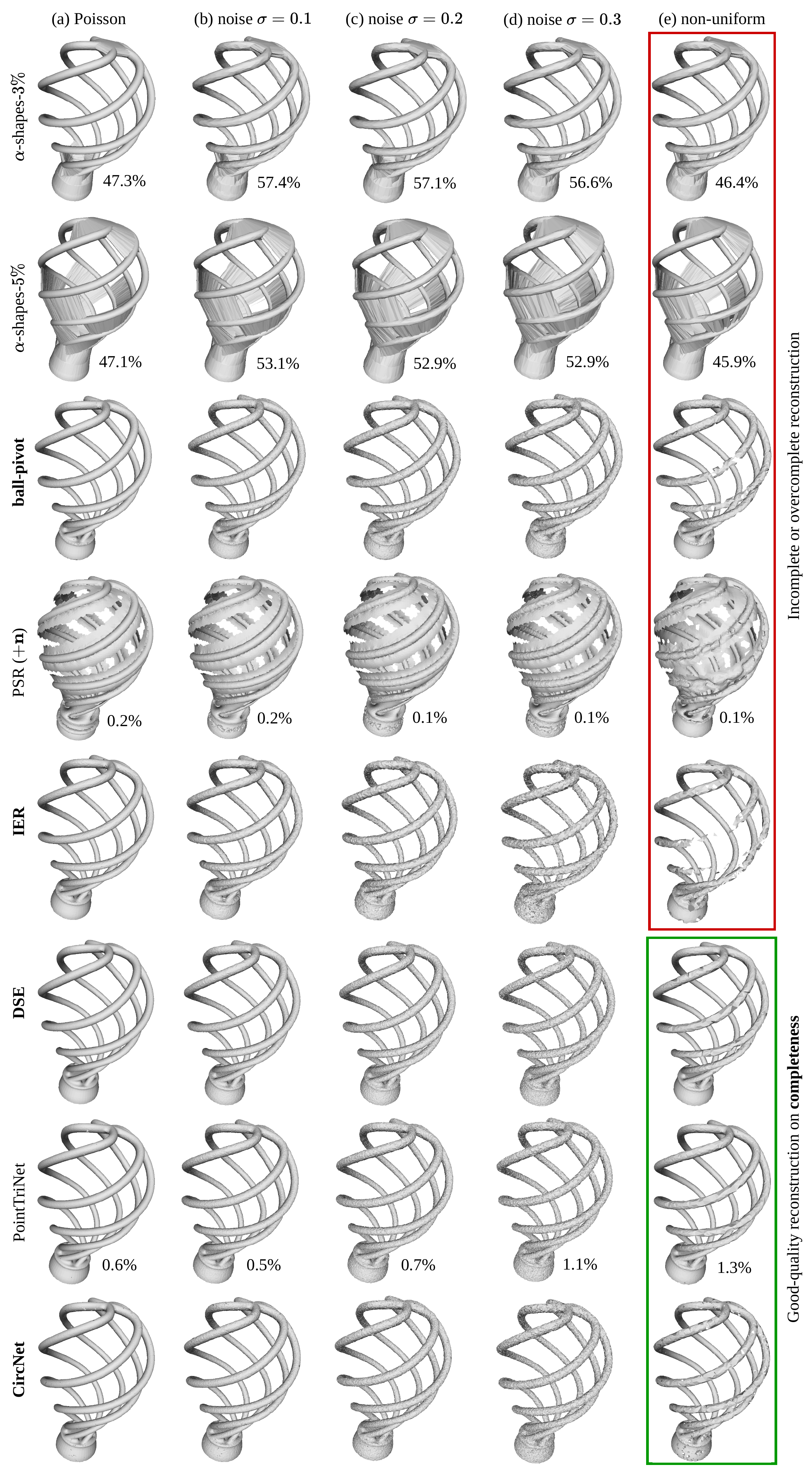}
    \vspace{-4mm}
    \caption{\color{black}Further examples of the reconstructed surface shapes of different methods for noisy and non-uniform point clouds.}
    \label{fig:more_vis_shape2}
\end{figure}

{\color{black}{\bf More visualizations.} Figure~\ref{fig:Mat3D_scene2} shows the reconstructed meshes of different methods for another building in Matterport3D. The proposed CircNet consistently produces mesh of high quality. We also compare the reconstruction quality of different methods, using complex shapes in the Thingi10K dataset~(\citealt{zhou2016thingi10k}). It can be seen from Fig.~\ref{fig:more_vis_shape1} that the learning-based triangulation methods, \ie~DSE, PointTriNet and the proposed CircNet are all robust to noise. For the non-uniform point cloud, only PointTriNet and the proposed CircNet reconstruct the complete underlying shape without chopping out the 8 thin tails. The other methods either reconstruct an incomplete mesh or overcomplete mesh. The number right below each mesh indicates the percentage of non-manifold edges of each specific mesh. Figure \ref{fig:more_vis_shape2} shows a further example.}

{\color{black}{\bf Hyperparameters.} We evaluate the effect of neighborhood size ($K$) and the number of predictions per anchor cell ($s$) on our model performance. Specifically, we compare the model performance of $K=50$ to that of $K\in\{25,100,200\}$. 
It can be seen from Table~\ref{table:hyper_params} that $K=100$ and $K=200$ perform similarly to $K=50$, while $K=25$ results in performance drop. The reason is that $K=25$ is too small to cover the 1-ring neighborhood. Given similar performance, smaller $K$ is preferred for better efficiency. 
As for $s$, we do not recommend settings of $s>2$ because dominantly, each anchor cell is observed to be populated by at most two ground-truth circumcenters. We hence compare the model performance of $s=2$ to that of $s=1$ only. It shows that $s=2$ performs slightly better on the overall reconstruction quality.}

\begin{table}[h]
\centering
    \caption{Poisson.}
    \label{table:uniform}
    \begin{adjustbox}{width=1\textwidth}
    {
    \begin{tabular}{l|c|c|c|c|c|c|c}
    \hline
    \multirow{3}{*}{Method}& \multicolumn{7}{c}{Surface Quality} \\
    \cline{2-8}
    & \multicolumn{5}{c|}{overall} & \multicolumn{2}{c}{sharp}\\
    \cline{2-8}
    & CD1($\times10^2$)$\downarrow$& CD2($\times10^5$)$\downarrow$ & F1$\uparrow$ & NC$\uparrow$ & NR$\downarrow$ &  ECD1($\times10^2$)$\downarrow$ & EF1$\uparrow$\\
    \hline
    $\alpha$-shapes-3\%&0.421 &2.460 &0.853 &0.948 &7.153 &4.067 &0.434 \\ 
    $\alpha$-shapes-5\%&0.536 &5.844 &0.834 &0.942 &7.650 &4.462 &0.442 \\
     ball-pivot (+$\mathbf{n}$)&0.286 &0.592 &0.938 &0.975 &4.279 &1.514 &0.607 \\
     \cline{1-8}
      DSE&0.276 &0.524 &0.952 &0.981 &3.560 &1.124 &0.667 \\
     IER &0.282 &0.572 &0.945 &0.981 &3.300 &1.065 &0.718 \\
    PointTriNet&0.278 &0.532 &0.950 &0.978 &3.921 &1.372 &0.667 \\
      \cline{1-8}
     CircNet~(Prop.)&  0.278 &0.533 &0.949 &0.976 &4.033&1.281&0.660 \\
     \hline
    \end{tabular}
    }
    \end{adjustbox}
    
    \centering
    \caption{Noise level: $\sigma=0.1$.}
    \label{table:noise_sigma_01}
    \begin{adjustbox}{width=1\textwidth}
    {
    \begin{tabular}{l|c|c|c|c|c|c|c}
    \hline
    \multirow{3}{*}{Method}& \multicolumn{7}{c}{Surface Quality}  \\
    \cline{2-8}
    & \multicolumn{5}{c|}{overall} & \multicolumn{2}{c}{sharp} \\
    \cline{2-8}
    & CD1($\times10^2$)$\downarrow$& CD2($\times10^5$)$\downarrow$ & F1$\uparrow$ & NC$\uparrow$ & NR$\downarrow$ &  ECD1($\times10^2$)$\downarrow$ & EF1$\uparrow$\\
    \hline
    $\alpha$-shapes-3\%&0.453 &2.501 &0.840 &0.948 &9.697 &4.579 &0.398\\ 
    $\alpha$-shapes-5\%&0.564 &5.746 &0.822 &0.943 &9.465 &5.301 &0.397\\
     ball-pivot (+$\mathbf{n}$)&0.339 &0.807 &0.892 &0.963 &10.260 &2.392 &0.544\\
     \cline{1-8}
     DSE&0.326 &0.711 &0.910 &0.970 &9.541 &1.187 &0.639 \\
     IER &0.348 &0.849 &0.881 &0.967 &9.242 &1.413 &0.655 \\
     PointTriNet&0.328 &0.718 &0.908 &0.966 &9.899 &1.581 &0.634 \\
      \cline{1-8}
     CircNet (Prop.)& 0.328 &0.720 &0.908 &0.965 &10.053 &1.500 &0.630\\ 
     \hline
    \end{tabular}
    }
    \end{adjustbox}
    
    \centering
    \caption{Noise level: $\sigma=0.2$.}
    \label{table:noise_sigma_02}
    \begin{adjustbox}{width=1\textwidth}
    {
    \begin{tabular}{l|c|c|c|c|c|c|c}
    \hline
    \multirow{3}{*}{Method}& \multicolumn{7}{c}{Surface Quality} \\
    \cline{2-8}
    & \multicolumn{5}{c|}{overall} & \multicolumn{2}{c}{sharp} \\
    \cline{2-8}
    & CD1($\times10^2$)$\downarrow$& CD2($\times10^5$)$\downarrow$ & F1$\uparrow$ & NC$\uparrow$ & NR$\downarrow$ &  ECD1($\times10^2$)$\downarrow$ & EF1$\uparrow$\\
    \hline
    $\alpha$-shapes-3\%&0.542 &3.031 &0.745 &0.942 &11.667 &6.034 &0.315\\ 
    $\alpha$-shapes-5\%&0.676 &6.896 &0.716 &0.938 &11.136 &6.918 &0.300 \\
     ball-pivot (+$\mathbf{n}$)&0.429 &1.347 &0.783 &0.931 &16.574 &4.901 &0.390 \\
     \cline{1-8}
     DSE&0.415 &1.226 &0.797 &0.937 &16.234 &2.695 &0.515  \\
     IER &0.446 &1.439 &0.761 &0.945 &13.702 &3.327 &0.500 \\
     PointTriNet&0.416 &1.234 &0.795 &0.934 &16.430 &3.519 &0.500  \\
      \cline{1-8}
      CircNet (Prop.)& 0.419 &1.245 &0.793 &0.931 &16.735 &4.396 &0.467 \\
     \hline
    \end{tabular}
    }
    \end{adjustbox}
    
    \centering
    \caption{Noise level: $\sigma=0.3$.}
    \label{table:noise_sigma_03}
    \begin{adjustbox}{width=1\textwidth}
    {
    \begin{tabular}{l|c|c|c|c|c|c|c}
    \hline
    \multirow{3}{*}{Method}& \multicolumn{7}{c}{Surface Quality}\\
    \cline{2-8}
    & \multicolumn{5}{c|}{overall} & \multicolumn{2}{c}{sharp} \\
    \cline{2-8}
    & CD1($\times10^2$)$\downarrow$& CD2($\times10^5$)$\downarrow$ & F1$\uparrow$ & NC$\uparrow$ & NR$\downarrow$ &  ECD1($\times10^2$)$\downarrow$ & EF1$\uparrow$\\
    \hline
    $\alpha$-shapes-3\%&0.653 &3.965 &0.622 &0.934 &13.471 &6.764 &0.260\\ 
    $\alpha$-shapes-5\%&0.798 &8.036 &0.562 &0.932 &12.529 &7.402 &0.245 \\
     ball-pivot (+$\mathbf{n}$)&0.524 &2.127& 0.697&0.898 &21.351 &6.819 &0.293 \\
     \cline{1-8}
     DSE&0.517 &2.033 &0.697 &0.883 &23.386 &6.311 &0.325 \\
     IER &0.545 &2.235 &0.668 &0.929 &16.030 &5.682 &0.340  \\
     PointTriNet&0.517 &2.038 &0.698 &0.888 &22.738 &6.353 &0.324  \\
      \cline{1-8}
      CircNet (Prop.)& 0.523 &2.076 &0.689 &0.880 &23.287 &6.790 &0.294 \\
     \hline
    \end{tabular}
    }
    \end{adjustbox}
    
     \centering
    \caption{Non-uniform data.}
    \label{table:non_uniform}
    \begin{adjustbox}{width=1\textwidth}
    {
    \begin{tabular}{l|c|c|c|c|c|c|c}
    \hline
    \multirow{3}{*}{Method}& \multicolumn{7}{c}{Surface Quality}  \\
    \cline{2-8}
    & \multicolumn{5}{c|}{overall} & \multicolumn{2}{c}{sharp} \\
    \cline{2-8}
    & CD1($\times10^2$)$\downarrow$& CD2($\times10^5$)$\downarrow$ & F1$\uparrow$ & NC$\uparrow$ & NR$\downarrow$ &  ECD1($\times10^2$)$\downarrow$ & EF1$\uparrow$\\
    \hline
    $\alpha$-shapes-3\%&0.475 &3.407 &0.825 &0.936 &8.875 &4.760 &0.367 \\ 
    $\alpha$-shapes-5\%&0.543 &5.723 &0.820 &0.939 &8.308 &4.987 &0.383 \\
     ball-pivot (+$\mathbf{n}$)&1.237 &43.581 &0.653 &0.929 &9.288 &8.538 &0.418 \\
     \cline{1-8}
     DSE&0.353 &1.325 &0.891 &0.955 &7.238 &2.842 &0.495 \\
     IER &0.542 &6.629 &0.833 &0.960 &6.044 &4.980 &0.547 \\
     PointTriNet&0.342 &1.103 &0.894 &0.956 &7.357 &2.535 &0.514 \\
      \cline{1-8}
     CircNet (Prop.) & 0.395 & 1.852 & 0.858 & 0.942 & 8.693 & 3.490 & 0.457 \\
     \hline
    \end{tabular}
    }
    \end{adjustbox}
\end{table}

\begin{table}[t]
 \centering
    \caption{Random.}
    \label{table:random}
    \begin{adjustbox}{width=1\textwidth}
    {
    \begin{tabular}{l|c|c|c|c|c|c|c}
    \hline
    \multirow{3}{*}{Method}& \multicolumn{7}{c}{Surface Quality}  \\
    \cline{2-8}
    & \multicolumn{5}{c|}{overall} & \multicolumn{2}{c}{sharp} \\
    \cline{2-8}
    & CD1($\times10^2$)$\downarrow$& CD2($\times10^5$)$\downarrow$ & F1$\uparrow$ & NC$\uparrow$ & NR$\downarrow$ &  ECD1($\times10^2$)$\downarrow$ & EF1$\uparrow$\\
    \hline
    $\alpha$-shapes-3\%&0.424 &2.514 &0.851 &0.946 &7.358 &4.248 &0.380 \\ 
    $\alpha$-shapes-5\%&0.533 &5.750 &0.834 &0.942 &7.634 &4.969 &0.392 \\
     ball-pivot (+$\mathbf{n}$)&0.356 &1.327 &0.871 &0.967 &5.471 &1.987 &0.554 \\
     \cline{1-8}
     PointTriNet&0.284 &0.565 &0.942 &0.973 &4.752 &1.863 &0.604 \\
     IER &0.289 &0.599 &0.940 &0.974 &4.475 &1.433 &0.640 \\
     DSE&0.282 &0.555 &0.945 &0.976 &4.254 &1.380 &0.601 \\
      \cline{1-8}
     CircNet (Prop.)& 0.316 &0.815 &0.905 &0.963 &5.699 &2.373 &0.553 \\
     \hline
    \end{tabular}
    }
    \end{adjustbox}   
\end{table}

\begin{table}[t]
 \centering
    \caption{\color{black}Effects of different hyperparameters on the performance.}
    \label{table:hyper_params}
    \begin{adjustbox}{width=1\textwidth}
    {
    \begin{tabular}{l|c|c|c|c|c|c|c}
    \hline
    \multirow{3}{*}{Method}& \multicolumn{7}{c}{Surface Quality}  \\
    \cline{2-8}
    & \multicolumn{5}{c|}{overall} & \multicolumn{2}{c}{sharp} \\
    \cline{2-8}
    & CD1($\times10^2$)$\downarrow$& CD2($\times10^5$)$\downarrow$ & F1$\uparrow$ & NC$\uparrow$ & NR$\downarrow$ &  ECD1($\times10^2$)$\downarrow$ & EF1$\uparrow$\\
    \hline
    $k=25, s=2$&0.285 &0.548 &0.949 &0.983 &2.012 &0.786 &0.917 \\ 
    $k=50, s=1$&0.285 &0.636 &0.950 &0.985 &1.768 &\textbf{0.686} &0.925 \\
    $k=50, s=2$&0.284 &0.545 &0.950 &0.985 &1.778 &0.712 &0.924 \\
    $k=100, s=2$&\textbf{0.284} &\textbf{0.544} &\textbf{0.950} &\textbf{0.985} &\textbf{1.766} &0.691 &\textbf{0.925} \\
    $k=200, s=2$&0.284 &0.545 &0.950 &0.985 &1.776 &0.697 &0.925 \\
     \hline
    \end{tabular}
    }
    \end{adjustbox}   
\end{table}

\end{document}